\documentclass[10pt,twocolumn,letterpaper]{article}

\usepackage[pagenumbers]{cvpr} % To force page numbers, e.g. for an arXiv version

\usepackage{times}
\usepackage{epsfig}
\usepackage{graphicx}
\usepackage{amsmath}
\usepackage{amssymb}
\usepackage{color}
\usepackage{algorithm}
\usepackage{algorithmic}
\usepackage{multirow}
\usepackage{booktabs}
\usepackage{pifont}
\usepackage{listings}
\usepackage{cite} % automatically sorts the citations
\usepackage{soul}

\usepackage{tabulary,overpic,xcolor}
\definecolor{citecolor}{RGB}{34,139,34}

\usepackage[pagebackref,breaklinks,colorlinks]{hyperref}

\newcommand{\cmark}{\ding{51}}
\newcommand{\xmark}{\ding{55}}

\newcommand{\app}{\raise.17ex\hbox{$\scriptstyle\sim$}}

\newcolumntype{x}[1]{>{\centering\arraybackslash}p{#1pt}}

\newlength\savewidth

\makeatletter\renewcommand\paragraph{\@startsection{paragraph}{4}{\z@}
  {.5em \@plus1ex \@minus.2ex}{-.5em}{\normalfont\normalsize\bfseries}}\makeatother

\usepackage[capitalize]{cleveref}
\crefname{section}{Sec.}{Secs.}
\Crefname{section}{Section}{Sections}
\Crefname{table}{Table}{Tables}
\crefname{table}{Tab.}{Tabs.}

\begin{document}

\title{X-DETR: A Versatile Architecture for Instance-wise Vision-Language Tasks}

\author{Zhaowei Cai \ \ Gukyeong Kwon \ \ Avinash Ravichandran \ \ Erhan Bas \\ Zhuowen Tu \ \ Rahul Bhotika \ \ Stefano Soatto\\
AWS AI Labs\\
{\tt\small \{zhaoweic,gukyeong,ravinash,erhanbas,ztu,bhotikar,soattos\}@amazon.com}
}

\maketitle

%%%%%%%%% ABSTRACT
\begin{abstract}
In this paper, we study the challenging instance-wise vision-language tasks, where the free-form language is required to align with the objects instead of the whole image. To address these tasks, we propose X-DETR, whose architecture has three major components: an object detector, a language encoder, and vision-language alignment. The vision and language streams are independent until the end and they are aligned using an efficient dot-product operation. The whole network is trained end-to-end, such that the detector is optimized for the vision-language tasks instead of an off-the-shelf component. To overcome the limited size of paired object-language annotations, we leverage other weak types of supervision to expand the knowledge coverage. This simple yet effective architecture of X-DETR shows good accuracy and fast speeds for multiple instance-wise vision-language tasks, e.g., 16.4 AP on LVIS detection of 1.2K categories at $\sim$20 frames per second without using any LVIS annotation during training.
\end{abstract}

%%%%%%%%% BODY TEXT
%=======================================================================
\section{Introduction}
\label{sec:intro}

\begin{figure*}[!t]
\centering
\centerline{\epsfig{figure=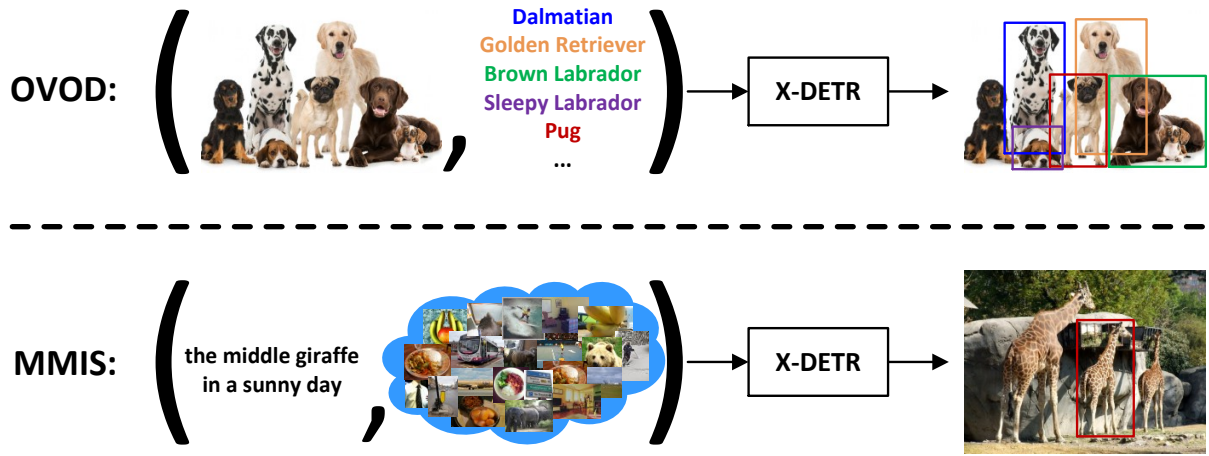,width=16cm,height=6.11cm}}
\caption{The illustration for open-vocabulary object detection (OVOD) and multi-modal instance search (MMIS).}
\label{fig:teaser}\vspace{-3mm}
\end{figure*}

Vision-language (V+L) understanding has achieved promising progresses in the past a few years \cite{DBLP:conf/nips/LuBPL19,tan2019lxmert,li2019visualbert,lu202012,DBLP:conf/eccv/ChenLYK0G0020,DBLP:conf/eccv/Li0LZHZWH0WCG20,desai2021virtex,DBLP:conf/icml/RadfordKHRGASAM21,DBLP:conf/icml/JiaYXCPPLSLD21}. \cite{DBLP:conf/icml/RadfordKHRGASAM21,DBLP:conf/icml/JiaYXCPPLSLD21} have shown that strong vision-language alignment can be enabled by a simple dot-product between vision and language representations, with the help of large-scale image-caption pairs (hundred of millions to billions). Although they have achieved very exciting results on image-level tasks, such as open-vocabulary classification and image-text retrieval, how to develop a system for instance-wise localization based V+L tasks is still unknown, e.g., open-vocabulary object detection (OVOD) and multi-modal instance search (MMIS). OVOD detects any object categories defined by free-form language descriptions without finetuning (see Figure \ref{fig:teaser} top), where the size of categories could span from dozens to thousands. On the other hand, MMIS  retrieves the most similar object region from a database given a free-form language query (see Figure \ref{fig:teaser} bottom), where the database size could be millions or billions for a commercial search engine.

%%%%%%%%%%%%%%%%%%%%%%%%% teaser %%%%%%%%%%%%%%%%%%%%%%%%%%%
% 513x193

One straightforward solution is to use a R-CNN framework \cite{DBLP:conf/cvpr/GirshickDDM14} with a pretrained object proposal detector, e.g., \cite{ren2015faster,cai2018cascade,DBLP:conf/eccv/CarionMSUKZ20,DBLP:conf/iclr/ZhuSLLWD21} and a pretrained V+L model, e.g., CLIP \cite{DBLP:conf/icml/RadfordKHRGASAM21}, denoted as R-CLIP. Then the pipeline is to 1) detect object proposals, and 2) crop image regions of proposals, and 3) forward the cropped image regions through the V+L model for final vision-language alignment. However, this framework has its limitations. First, it is very slow because feature extraction is repeated in overlapped image regions. Speed is the key for applications of instance-wise V+L tasks. For example, OVOD usually requires real-time speeds  (e.g., 25 frames per second),
and MMIS requires instant retrieval results (e.g., in at most a few seconds) from a database of millions or billions instances. Second, the image-level representation of CLIP is suboptimal for the instance-wise V+L tasks. For example, the cropped regional representation lacks global context, which is required in some tasks. As shown in the example of locating ``the middle giraffe in a sunny day'' in Figure \ref{fig:teaser} (bottom), the conditions of ``middle'' and ``in a sunny day'' require to know the relations with other giraffes and  the global image context, respectively.

To resolve these issues, we propose an efficient and effective architecture for various instance-wise V+L tasks, denoted as X-DETR (cross-modal DETR). 
It has three major components: a visual object detector (transformer based DETR \cite{DBLP:conf/eccv/CarionMSUKZ20}), a language encoder (a RoBERTa model \cite{liu2019roberta}) and alignment between the visual instance and the language description. This model is trained end-to-end, optimizing all components simultaneously.
Hence, the detector is adapted to the instance-wise V+L tasks instead of as an off-the-shelf component as in the previous V+L efforts \cite{DBLP:conf/nips/LuBPL19,DBLP:conf/eccv/ChenLYK0G0020,DBLP:conf/eccv/Li0LZHZWH0WCG20}. Furthermore, motivated by the success of CLIP \cite{DBLP:conf/icml/RadfordKHRGASAM21}, we keep the visual and language streams independent as much as possible, and align them together by a simple dot-product operation at the very end, instead of using an expensive joint-modality transformer as in \cite{DBLP:conf/nips/LuBPL19,DBLP:conf/eccv/ChenLYK0G0020,DBLP:conf/eccv/Li0LZHZWH0WCG20,kamath2021mdetr}. In this sense, X-DETR can be seen as a detection counterpart of CLIP, but it overcomes the two issues of R-CLIP framework. First, thanks to the architecture design of X-DETR, the representations of multiple instances and language queries can be obtained by a single feed-forward pass. For example, X-DETR can run 20 fps for OVOD on LVIS of $\sim$1.2K categories \cite{gupta2019lvis}, and for MMIS, it can retrieve the object in seconds from one million of instances given a query. Second, each instance feature representation of DETR encodes the information from global image and thus can be more accurate for tasks requiring global context. For example, X-DETR is much better than R-CLIP on RefCOCO/RefCOCO+ datasets \cite{DBLP:conf/eccv/YuPYBB16} requiring context information. 

A cornerstone of the success of CLIP is the large-scale training data, $\sim$400 million image-caption pairs. The paired image-caption annotation is relatively easy to collect, e.g., by crawling from the internet. However, paired object-language data usually requires human annotation and thus is very expensive. As a result, only a few datasets have this kind of annotations, i.e., Flickr30k entities \cite{DBLP:journals/ijcv/PlummerWCCHL17} for grounding, RefCOCO/RefCOCO+/RefCOCOg \cite{DBLP:conf/eccv/YuPYBB16,DBLP:conf/cvpr/MaoHTCY016} for referring expression comprehension (REC), VG \cite{DBLP:journals/ijcv/KrishnaZGJHKCKL17} for dense captioning and GQA \cite{hudson2019gqa} for visual question answering (VQA). In fact, the union of them is of relatively small size, with only $\sim$90K unique images for training, which is not enough to learn universal instance-wise V+L representations. To deal with this challenge, our framework resorts to other types of weak supervisions than the paired object-language annotation, including image-caption pairs, object bounding boxes and pseudo-labels, from datasets including COCO \cite{DBLP:conf/eccv/LinMBHPRDZ14}, OpenImages \cite{krasin2017openimages}, GCC \cite{DBLP:conf/acl/SoricutDSG18}, and LocNar \cite{DBLP:conf/eccv/Pont-TusetUCSF20}. This expanded combination of full and weak annotations provides broader knowledge coverage for X-DETR to learn universal representation, as will be seen in our experiments. 

Our contributions can be summarized as follows:
\begin{itemize}
    \item We propose a simple yet effective architecture, X-DETR, which is end-to-end optimized for various instance-wise V+L tasks, such as OVOD, MMIS, phrase grounding, and referring expression. It also shows better transferring capacity on downstream detection tasks than other detectors.
    
    \item We have empirically shown that the CLIP-style of vision-language alignment, i.e., simple dot-product, can achieve good results with fast speeds for instance-wise V+L tasks, and the expensive cross-modality attention may not be necessary.
    
    \item We have shown that X-DETR is capable of using different weak supervisions, which are helpful to expand the knowledge coverage of the model.
\end{itemize}

%=======================================================================
\section{Related Work}

\paragraph{Vision-Language Learning}
is a popular interdisciplinary research topic \cite{DBLP:conf/nips/LuBPL19,tan2019lxmert,li2019visualbert,lu202012,DBLP:conf/eccv/ChenLYK0G0020,DBLP:conf/eccv/Li0LZHZWH0WCG20,desai2021virtex,DBLP:conf/icml/RadfordKHRGASAM21}. Early efforts focused on a single specific task, e.g., \cite{mao2014deep,xu2015show} on image captioning, \cite{anderson2018bottom,yang2016stacked} on visual question answering (VQA) \cite{antol2015vqa}, \cite{lee2018stacked,miech2021thinking} on image-text retrieval and \cite{hu2016natural,yu2018mattnet} on REC \cite{kazemzadeh2014referitgame}, etc.  Recent efforts  have focused on the joint pretraining of two modalities \cite{DBLP:conf/nips/LuBPL19,tan2019lxmert,li2019visualbert,lu202012,DBLP:conf/eccv/ChenLYK0G0020,DBLP:conf/eccv/Li0LZHZWH0WCG20}, aiming for a multi-task model that can work on multiple downstream tasks simultaneously. Although these methods do not necessarily work on detection related tasks, most of them use an off-the-shelf detector, e.g., \cite{anderson2018bottom,zhang2021vinvl}, for more accurate visual feature representation. However, the system is not end-to-end and it is not guaranteed that the detector is optimized for the following V+L tasks. Differently, X-DETR, a versatile architecture for multiple instance-wise V+L tasks, has all components end-to-end optimized.

%%%%%%%%%%%%%%%%%%%%%%%%% framework %%%%%%%%%%%%%%%%%%%%%%%%%%%
% 467x201
\begin{figure*}[!t]
\centering
\centerline{\epsfig{figure=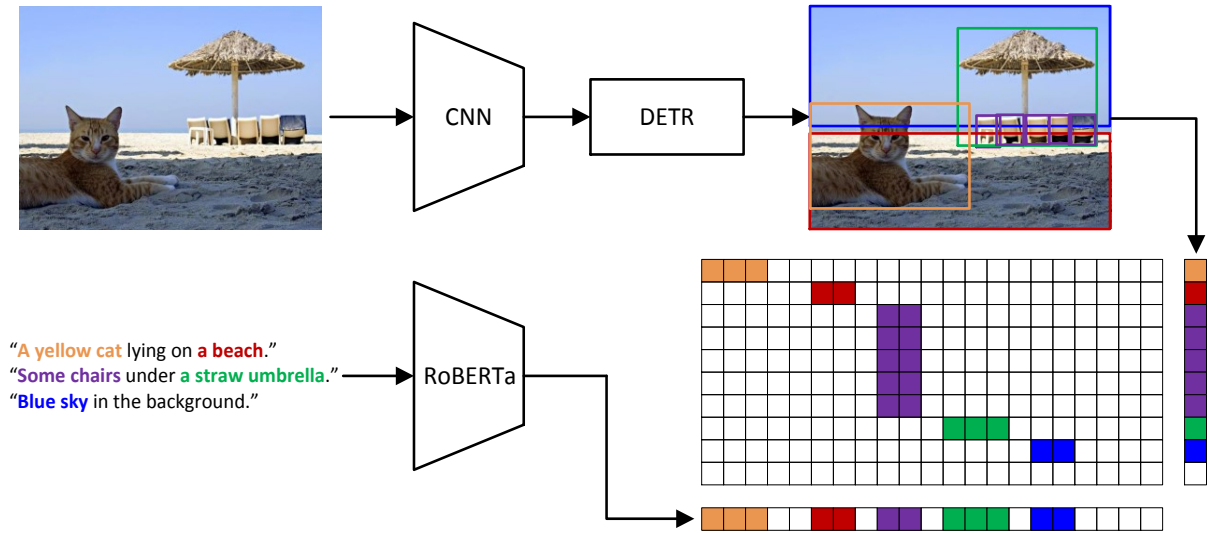,width=16cm,height=7.08cm}}
\caption{Overview of our X-DETR architecture. The same color of text and object means they are aligned concepts. The matrix is the similarity matrix between text tokens and instance hypotheses, where the colored score items should be maximized but the rest should be minimized. }
\label{fig:framework}\vspace{-3mm}
\end{figure*}

\paragraph{Uni-modal Object Detection}
has achieved great progresses in recent years, including the pioneering two-stage object detectors \cite{DBLP:conf/cvpr/GirshickDDM14,ren2015faster,lin2017feature,cai2018cascade},  efficient one-stage detectors \cite{redmon2016you,liu2016ssd}, and the very recent transformer-based object detectors \cite{DBLP:conf/eccv/CarionMSUKZ20,DBLP:conf/iclr/ZhuSLLWD21}. However, these uni-modal detectors are constrained to predefined categories, e.g., COCO/Objects365/OpenImages \cite{DBLP:conf/eccv/LinMBHPRDZ14,shao2019objects365,krasin2017openimages} of 80/365/601 classes, and they are unable to detect any  categories beyond the predefined ones. In natural images, the objects are orders more diverse than the predefined categories (usually 10$\sim$1,000) in the sense of categories, attributes (e.g., colors, materials, etc.), geometric location (e.g., ``man on the right''), relation to the environment (``man sitting in a couch''), etc. 
The traditional uni-modal object detectors are incapable of dealing with these problems.

\paragraph{Multi-modal Object Detection}
tries to detect objects with free-form language description.  Previous work \cite{hinami2017discriminative,yang2019fast,plummer2020revisiting} extended traditional uni-model detection framework to accommodate the language inputs. However, since those efforts are usually not end-to-end, they do not leverage the data from the two modalities very well and are not performant.
Recently, MDETR \cite{kamath2021mdetr}, a modulated detector for multi-modal understanding, achieved  state-of-the-art results on multiple datasets, including phrase grounding \cite{DBLP:journals/ijcv/PlummerWCCHL17}, REC \cite{DBLP:conf/eccv/YuPYBB16,DBLP:conf/cvpr/MaoHTCY016}, GQA \cite{hudson2019gqa}, etc.
However,  it adopts the expensive joint-modality transformer to modeling the alignment between language and vision, which prevents it from being used in practical applications like OVOD and MMIS. For example, it takes 5 seconds per image for OVOD and hours for MMIS.
OVOD \cite{hinami2017discriminative,plummer2020revisiting,zareian2021open}  usually leverages language models as it is impossible to pre-define all open-vocabulary categories. It is related to zero-shot object detection (ZSOD) \cite{bansal2018zero,rahman2018zero}, where the model is trained on the seen categories and evaluated on unseen ones. The language model is usually used to define the semantic space of the seen and unseen categories. 
X-DETR falls into the category of OVOD because it is trained on large V+L data. 

\vspace{-1mm}
\paragraph{Cross-modal Retrieval}
Cross-modal (image and text) retrieval has a long  research history\cite{rasiwasia2010new}. 
This task is also popular in recent V+L learning efforts \cite{miech2021thinking,wang2018learning}. Many X-DETR tasks are related to cross-modal retrieval, but they are instance-wise. For example, OVOD \cite{zareian2021open,hinami2017discriminative,plummer2020revisiting} and MMIS \cite{hinami2017discriminative,plummer2020revisiting,liu2021ovis} are to retrieve the most similar bounding boxes across the full dataset given the free-form query, and phrase grounding \cite{DBLP:journals/ijcv/PlummerWCCHL17} and REC \cite{DBLP:conf/eccv/YuPYBB16,DBLP:conf/cvpr/MaoHTCY016} are to retrieve the most similar bounding box in an image given a query. OVOD and MMIS are more challenging tasks than grounding and REC because they have orders larger negative pool and not well explored yet. X-DETR aims to tackle them in an efficient and effective manner.

%%%%%%%%%%%%%%%%%%%%%%%%% architecture %%%%%%%%%%%%%%%%%%%%%%%%%%%
\begin{figure*}[t]
\begin{minipage}[b]{.64\linewidth}
\centering
%376x179
\centerline{\epsfig{figure=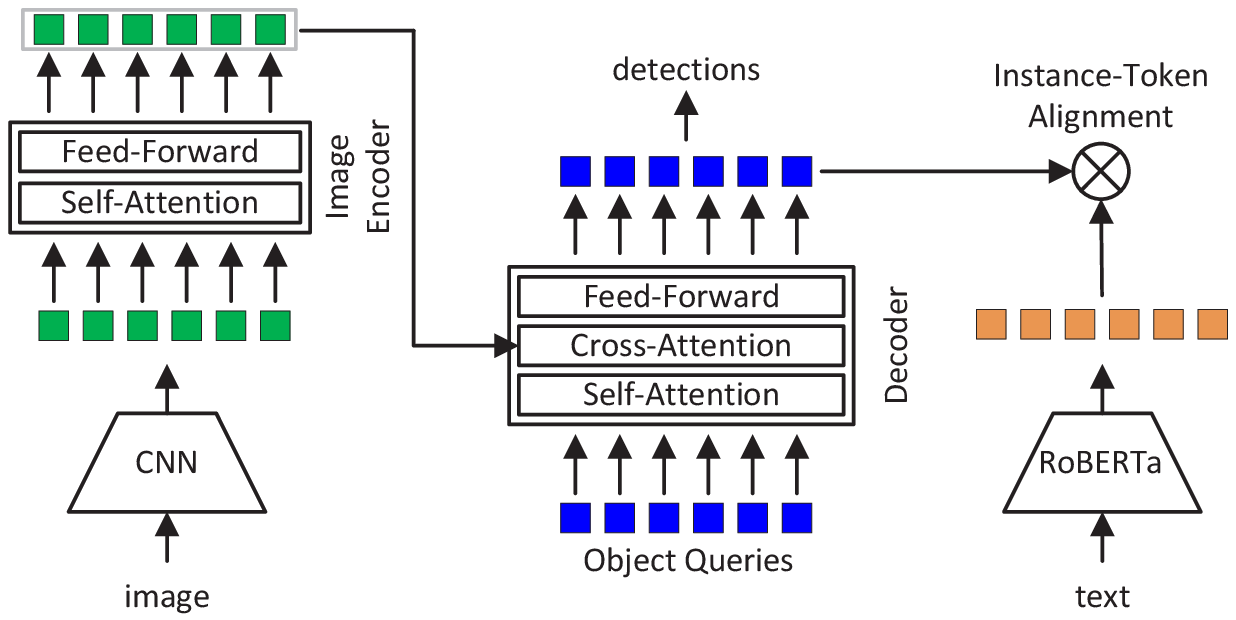,width=9.85cm,height=4.69cm}}{(a)}
\end{minipage}
\hfill
\begin{minipage}[b]{.35\linewidth}
\centering
%204x143
%\vspace{-5mm}
\centerline{\vspace{3mm}\epsfig{figure=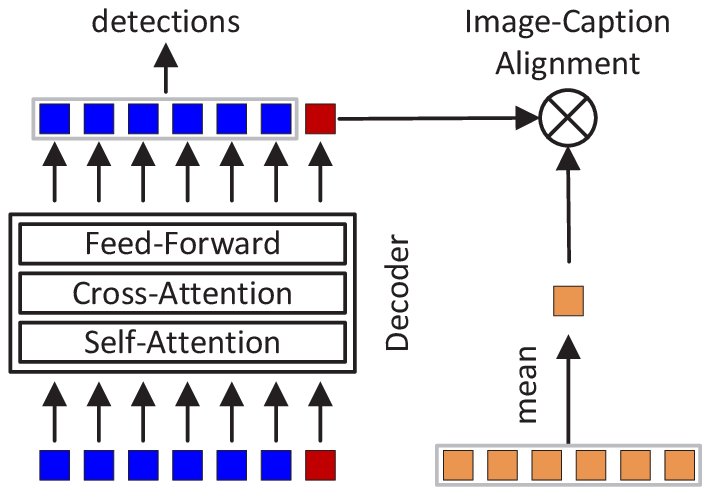,width=5.35cm,height=3.75cm}}(b)
\end{minipage}
\caption{(a) is the architecture of X-DETR. (b) illustrates how the encoded image-level query (red square) is aligned with the caption (yellow square).
}
\label{fig:xdetr}\vspace{-3mm}
\end{figure*}

%=======================================================================
\section{Instance-wise Cross-modality Network}
\label{sec:xdetr}

%------------------------------------------------------------------------
\subsection{Overall Architecture}
Figure \ref{fig:framework} gives an overview of X-DETR architecture with its three components: a visual object detector $D$, a text encoder $\psi$ and an alignment between visual instance and textual description $h$. The method takes an image $I$ and a language query $y$ as inputs and outputs the detected object $o$ and its alignment score with the language query $y$. The object detector $D$ is used to generate the instance $o=D(I)$ by processing the input image $I$, and the language encoder $\psi$ is to encode the tokenized text input $y$ into embeddings $\psi(y)$ which can be mapped to the joint space with the visual instance embeddings $o$. Vision-language alignment $h$ is the key component in V+L models, aligning the visual instance $o$ and language description $y$ in a joint feature space, such that $h(o,\psi(y))$ is higher (lower) for paired (unpaired) visual instances and language descriptions. For example, as shown in Figure \ref{fig:framework}, the alignment will pull the same concepts together, e.g., the detected cat (yellow bounding box) and the language description ``a yellow cat'', but push different concepts away, e.g., the cat and other language descriptions like ``a beach'', ``some chairs'', ``blue sky'', etc. In addition to the object-language alignment, X-DETR also has an image-language alignment component to leverage the weak data of image-caption pairs to learn broader knowledge coverage. The details of each component are discussed next.

%------------------------------------------------------------------------
\subsection{Object Detection}
\label{subsec:detection}

There are many choices for the object detection component \cite{lin2017feature,cai2018cascade,redmon2016you,liu2016ssd,DBLP:conf/eccv/CarionMSUKZ20,DBLP:conf/iclr/ZhuSLLWD21}, and we are not constrained to any specific one. We chose the transformer based framework of DETR \cite{DBLP:conf/eccv/CarionMSUKZ20}, because 1) it is simpler without heuristics compared with the other popular one or two-stage frameworks \cite{lin2017feature,cai2018cascade,redmon2016you,liu2016ssd}, and 2) each detected instance encodes information from global image due to the attention mechanism of transformers. In DETR, a standard CNN (e.g., ResNet \cite{DBLP:conf/cvpr/HeZRS16}) is applied to an image $I$ to extract convolutional feature maps. They are then flattened into a sequence of features passed through a encoder-detector transformer architecture \cite{DBLP:conf/nips/VaswaniSPUJGKP17}. To detect objects, the detector also takes object queries as input, and outputs the decoded queries as the detection results $o$. The detection architecture of X-DETR is shown in Figure \ref{fig:xdetr} (a). Due to the slow convergence of the original DETR, we resort to Deformable DETR \cite{DBLP:conf/iclr/ZhuSLLWD21}, which has faster convergence speed and better accuracy. Please refer to \cite{DBLP:conf/eccv/CarionMSUKZ20,DBLP:conf/iclr/ZhuSLLWD21} for more details.

The detector in X-DETR is a stand-alone component, where the detection results are conditioned on only the image input. This is different from previous multi-modal object detection frameworks \cite{yang2019fast,kamath2021mdetr}, where the detection results depend on both image and language inputs. Decoupling the vision and language streams makes the detection results independent of the queries. This is closer to a human detection system, where salient objects can be detected before being asked to detect something of interest.

%------------------------------------------------------------------------
\subsection{Object-Language Alignment}
\label{subsec:object-lanauge alignment}

A common and powerful strategy for image-level vision-language alignment is to leverage the interaction between two modalities by a joint-modality transformer \cite{DBLP:conf/nips/LuBPL19,DBLP:conf/eccv/ChenLYK0G0020,DBLP:conf/eccv/Li0LZHZWH0WCG20}, $MA$, as follows:
\begin{equation}
    h(I,y)=MA(\phi(I), \psi(y)),
\end{equation}
where $\phi$ is the image encoder, e.g., CNN. The joint-modality transformers could be either self-attention on concatenated vision and language features, e.g., \cite{DBLP:conf/eccv/ChenLYK0G0020,DBLP:conf/eccv/Li0LZHZWH0WCG20}, or cross-attention between vision and language streams, e.g., \cite{DBLP:conf/nips/LuBPL19}. 
Although more cross-modal interaction could lead to stronger cross-modal representations, the computation of $MA$ needs to be repeated if any vision or language input is changed. This is impractical for tasks such as OVOD and MMIS. For example, $MA$ needs to be repeated for 1.2K times for LVIS detection, 
and one million times for MMIS of database image size of one million, although $\phi(I)$ and $\psi(y)$ do not need to be recomputed.

Motivated by CLIP \cite{DBLP:conf/icml/RadfordKHRGASAM21}, where strong cross-modal alignment can be enabled by a simple and efficient dot-product operation, we adopt a similar approach, i.e., the vision-language alignment $h$ is a linear mapping of the two modality streams into a common feature space  followed by a dot-product operation between them,
\begin{equation}
\label{equ:dot product}
    h(o,y)=f(o) \odot g(\psi(y)),
\end{equation}
where $f$ and $g$ are linear mapping for instance representation $o$ and text representation $\psi(y)$, respectively, and $\odot$ is a dot-product operation. Since $o$ is generated by the stand-alone object detector $D$, (\ref{equ:dot product}) can be formulated as
\begin{equation}
    h(o,y)=f(D(I)) \odot g(\psi(y)).
\end{equation}
Since the two modalities are coupled at the very end via an efficient dot-product, no additional computation needs to be repeated. This is a key difference with methods that use joint-modality transformers \cite{DBLP:conf/nips/LuBPL19,DBLP:conf/eccv/ChenLYK0G0020,DBLP:conf/eccv/Li0LZHZWH0WCG20,kamath2021mdetr} and enables X-DETR to be used in practical tasks without sacrificing accuracy, e.g., real-time OVOD and instant MMIS.

%------------------------------------------------------------------------
\subsection{Image-Language Alignment}

As shown in Figure \ref{fig:xdetr} (b), in addition to the queries used for object detection, an additional query is added to the query list as image query, which encodes image-level representation instead of instance-level representation like the other object queries. All queries are forwarded to the decoder with no difference. The language representation is the mean of all token representations in the sentence. Similar to CLIP, the alignment is also a simple dot-product between the encoded image query and the caption representation.

%=======================================================================
\section{Training}
\label{sec:training}

In this section, we describe our multi-task loss design involving losses from object detection and vision-language alignment, and different types of data we used.

%------------------------------------------------------------------------
\subsection{Class-agnostic Object Detection}

In DETR \cite{DBLP:conf/eccv/CarionMSUKZ20}, object detection is a set-to-set prediction, where a set of queries is firstly mapped to a set of ground truth objects using Hungarian matching during training. 
After matching, the matched object query is learned to regress to the corresponding ground truth, with a classification loss (cross-entropy) and bounding box regression loss (generalized IoU and L1 loss). Different from the general object detection of \cite{DBLP:conf/eccv/CarionMSUKZ20,DBLP:conf/iclr/ZhuSLLWD21}, X-DETR does not use the class information of each object. Instead, it is class-agnostic detection: classifying a hypothesis to foreground or background. In total, three losses come from detection, a binary cross-entropy, a generalized IoU and L1 regression loss.

%------------------------------------------------------------------------
\subsection{Vision-Language Alignment}
\label{subsec:alignment loss}

We have explored different levels of vision-language alignment, including object-phrase, object-sentence and image-caption alignment. The diverse levels of alignments allow us to expand the training data, which is one of our key contributions. 

\vspace{-1mm}
\paragraph{Object-Phrase Alignment}
In the phrase grounding dataset of Flickr30k entities \cite{DBLP:journals/ijcv/PlummerWCCHL17}, the ground truth is a pair of sentence phrases and objects. For example in Figure \ref{fig:framework}, given the sentence ``Some chairs under a straw umbrella.'', the phrase queries are ``Some chairs'' and ``a straw umbrella'', with associated one or a few bounding boxes. We used the contrastive loss of InfoNCE \cite{oord2018representation} to optimize for the object-phrase alignment. The similarity of every potential object-token pair is computed, as shown in similarity matrix of Figure \ref{fig:framework}, and contrastive loss is applied for each row (object-token alignment) and column (token-object alignment) of this matrix. Note that although this loss does not directly optimize the phrase-object alignment, it achieves similar results. 

\vspace{-1mm}
\paragraph{Object-Sentence Alignment} is a special case of object-phrase alignment, where the length of the text is the full sentence, with data from REC datasets \cite{DBLP:conf/eccv/YuPYBB16,DBLP:conf/cvpr/MaoHTCY016}. Contrastive loss is applied between the object query and whole sentence embedding, which is averaged from token embeddings. 

\vspace{-1mm}
\paragraph{Image-Caption Alignment}
leverages the large-scale weak data of image-caption pairs. Similar to CLIP, the loss is a cross-modality contrastive loss between the encoded image queries and the captions.

%------------------------------------------------------------------------
\subsection{Training Efficiency}
\label{subsec:training efficiency}

In a typical image, there are multiple objects associated multiple language descriptions. The most efficient way for training is to use all object-language pairs at a single forward-backward pass. However, this is problematic for the models with cross-modality interaction $MA$, e.g., \cite{DBLP:conf/eccv/ChenLYK0G0020,DBLP:conf/eccv/Li0LZHZWH0WCG20,kamath2021mdetr}, due to the reasons discussed in Section \ref{subsec:object-lanauge alignment}. As a compromise, they usually merge all independent text queries into a paragraph as a single query, and then computes $MA$ only once. However, this violates the independence assumption of queries and queries can see each other during training, due to the property of transformer. At inference, only a single query is provided each time. 
On the contrary, X-DETR does not have these problems, due to the design of fully independent vision and language streams. As a benefit, X-DETR can take queries as many as possible at high efficiency while not violating the independence assumption, thus better performances.

%------------------------------------------------------------------------
\subsection{Training Data}

The most preferred data to learn instance-wise cross-modal representation is the paired object-language annotations, but they are very expensive and limited. X-DETR is capable of leveraging other weak supervision to cover broader knowledge.

\paragraph{Object-Language Data}
The paired object-language data comes from Flickr30k entities \cite{DBLP:journals/ijcv/PlummerWCCHL17} for grounding, RefCOCO/RefCOCO+/RefCOCOg \cite{DBLP:conf/eccv/YuPYBB16,DBLP:conf/cvpr/MaoHTCY016} for REC, VG \cite{DBLP:journals/ijcv/KrishnaZGJHKCKL17} for dense captioning and GQA \cite{hudson2019gqa} for VQA. We used the \emph{mixed} dataset of them following MDETR \cite{kamath2021mdetr}. Please refer to \cite{kamath2021mdetr} for more details.

\paragraph{Object Detection Data}
Since X-DETR has a stand-alone object detector, it can leverage data with detection annotation only. COCO \cite{DBLP:conf/eccv/LinMBHPRDZ14} bounding box annotations are used because many images of the \emph{mixed} dataset are from COCO. But category information is not used, since the detection in X-DETR is class-agnostic as described in Section \ref{subsec:detection}. Instead of having an image with detection annotations only, we add COCO objects into the existing images of the \emph{mixed} dataset, since many of them are sparsely annotated. For example, usually only 3-5 objects are annotated per image in REC datasets.

\paragraph{Image-Caption Data}
Many datasets have image-caption annotations, e.g., Flickr30 \cite{DBLP:journals/ijcv/PlummerWCCHL17}, COCO Captioning \cite{chen2015microsoft}, CC \cite{DBLP:conf/acl/SoricutDSG18}, SBU \cite{DBLP:conf/nips/OrdonezKB11}, etc. Since the \emph{mixed} dataset already includes Flickr30k and COCO images, we added their captions into the \emph{mixed} dataset. Beyond that, we also use the large-scale CC and Localized Narratives (LocNar) \cite{DBLP:conf/eccv/Pont-TusetUCSF20} (only the subset of OpenImages \cite{krasin2017openimages}). Note that we do not use the weak localization annotations of LocNar, i.e., the pointer tracks along narratives, because we found they are quite noisy and have no much benefit.

\begin{table*}[t]
\setlength{\tabcolsep}{6pt}
\begin{center}
\small
\begin{tabular}{lcccccccc} 
\toprule
Method & Data &Train Time &Test Time & AP & AP50 & AP$_\mathrm{r}$ & AP$_\mathrm{c}$ & AP$_\mathrm{f}$   \\\midrule
R-CLIP & 0\% &- &5s &12.7 & 19.3 & 17.0 & 16.0 & 9.0\\
R-CLIP+ & 0\% &- &10.6s &13.7 & 20.6 & 18.5 & 17.3 & 9.6\\
MDETR \cite{kamath2021mdetr} & 0\%  &- &5s & 6.4 & 9.1 & 1.9 & 3.6 &  9.8 \\
X-DETR (ours) & 0\%  &- &0.05s  & 16.4 & 24.4 & 9.6 & 15.2 & 18.8 \\
\midrule
DETR \cite{DBLP:conf/eccv/CarionMSUKZ20} & 1\% &0.5h &0.05s & 4.2 & 7.0 & 1.9 & 1.1 & 7.3 \\
MDETR \cite{kamath2021mdetr}  & 1\%  &11h &5s & 16.7 & 25.8 & 11.2 & 14.6 &  19.5 \\
X-DETR (ours) & 1\%  &1h &0.05s & 22.8 & 35.0 & 17.6 & 22.0 & 24.4 \\
\midrule
DETR \cite{DBLP:conf/eccv/CarionMSUKZ20} & 10\% &3h &0.05s &13.7 & 21.7 &4.1 & 13.2 & 15.9 \\
MDETR \cite{kamath2021mdetr}  & 10\%  &108h &5s & 24.2 & 38.0 &20.9 & 24.9 & 24.3 \\ 
X-DETR (ours) & 10\% &5.2h &0.05s & 29.5 & 44.7 &29.4 & 30.6 & 28.6 \\ 
\midrule
Mask R-CNN \cite{he2017mask} & 100\% &16h &0.1s &33.3 & 51.1 & 26.3 & 34.0 & 33.9\\
DETR \cite{DBLP:conf/eccv/CarionMSUKZ20} & 100\% &35h &0.05s & 17.8 & 27.5 & 3.2 & 12.9 & 24.8\\
MDETR \cite{kamath2021mdetr}  & 100\% &1080h &5s & 22.5 & 35.2 & 7.4 & 22.7 & 25.0 \\
X-DETR (ours) & 100\% &45h &0.05s & 34.0 & 49.0 & 24.7 & 34.6 & 35.1 \\
\bottomrule
\end{tabular}
\caption{OVOD detection results on LVIS-v1 (box AP). We followed \cite{kamath2021mdetr} to evaluate on the 5k \emph{minival} subset. The subscript ``r/c/f'' of AP is for rare/common/frequent categories of LVIS.}
\label{tab:lvis}
\end{center}\vspace{-5mm}
\end{table*}

\paragraph{Pseudo-Labeled Data}
We also used the pseudo labeled data generated by our X-DETR model on LocNar. Given an image and its corresponding caption, at first we use Spacy\footnote{https://spacy.io/} to extract the noun phrases which are possible objects in the corresponding image. Then we treat the pseudo-labeling as a phrase grounding task, retrieving the bounding box that is most aligned with the noun phrase. This pseudo-labeled dataset is then used with no difference to the other object-language pairs. In addition, the OpenImages object annotations were also added to LocNar similar to COCO.

%=======================================================================
\section{Experiments}

In this section, we are going to show how X-DETR performs on the challenging and practical OVOD and MMIS tasks, and it can be generalized well on the other simper tasks, such as phrase grounding and REC.

%------------------------------------------------------------------------
\subsection{Implementation Details}

In X-DETR, we used RoBERTa-base model as the text encoder (with implementation and pretrained model from HuggingFace \cite{wolf2019huggingface}), ResNet as the vision backbone, and Deformable DETR as the object detector. The model has been trained for 10 epochs, w.r.t. the \emph{mixed} dataset, with class-agnostic detection, object-phrase alignment, and image-caption alignment losses introduced in Section \ref{sec:training}. The object-sentence alignment loss is not used during pretraining but used in some of the downstream tasks finetuning. In each mini-batch, the batch size for fully/pseudo/weakly-annotated data is 4/2/4 for a single GPU. The image is resized such that the minimum of width and height is 600, which is smaller than the standard practice of 800 in object detection \cite{lin2017feature,he2017mask}, for training cost reduction. See the appendix for more training details. We report the results with and without finetuning. X-DETR without finetuning is a multi-task architecture for various instance-wise V+L tasks. All the compared algorithms use ResNet-101 (except R-CLIP). We do not use any prompt engineering as in CLIP \cite{DBLP:conf/icml/RadfordKHRGASAM21}, e.g., adding a prefix of ``a photo of'' to the language query.

In the baseline of R-CLIP with the pipeline introduced in Section \ref{sec:intro}, the proposal detector is a binary Faster R-CNN detector \cite{lin2017feature} of ResNet-50, trained on the COCO \emph{train2017} with binary (object/non-object) annotations. Note the binary detector is not a RPN network \cite{ren2015faster}. The V+L model is a pretrained ViT-B-32 CLIP model \cite{DBLP:conf/icml/RadfordKHRGASAM21}. For each image, the top 300 detection results are selected as the proposals. Since the proposals are usually tight bounding boxes, covering few context region. To include some context, we expand the proposal region by 50\% on each side of the bounding box. When using the context, the CLIP feature representations from the original and expanded regions are averaged before $L$2 normalization. This is denoted as R-CLIP+.

To evaluate the similarity between a free-form language query and an object is straightforward for R-CLIP and X-DETR. The whole sentence is forwarded through the language encoder to get feature embeddings for the input tokens. The feature embeddings corresponding to the phrase or the full sentence (excluding start and end tokens) are then averaged and $L$2-normalized as the phrase/sentence feature representation, which is then dot-producted with the $L$2-normalized object feature representation.

%------------------------------------------------------------------------
\subsection{Open-vocabulary Object Detection (OVOD)}

First, X-DETR is evaluated on OVOD task \cite{hinami2017discriminative,plummer2020revisiting,zareian2021open}. LVIS \cite{gupta2019lvis}, consisting of $\sim$1.2K categories, is used for OVOD evaluation. Two settings are evaluated: 1) without finetuning and 2) finetuning with different amount (i.e., 1\%/10\%/100\%) of LVIS annotations. The former is for OVOD and the latter is to show the transferring ability of X-DETR to other downstream detection datasets. For finetuning, the model is finetuned for 50 epochs, with the object detection and object-sentence alignment losses, learning rate dropped at 40th epoch and an image resolution of 800. For X-DETR and R-CLIP, the final score of an instance is the product between the objectness score and the classification probabilities over the categories of interest. 

The results are shown in Table \ref{tab:lvis}. For OVOD, R-CLIP achieves 12.7 AP even when the CLIP model is trained with image-caption annotations, and including some context information gives additional gains of 1 AP, but with 2 times slower speeds. X-DETR outperforms R-CLIP by 3.7 points and is about 100 times faster, without using any LVIS annotations. When compared with MDETR, the recent state-of-the-art localization based V+L model, X-DETR is 10 points better. Note that X-DETR is even close to the fully-supervised vanilla DETR baseline (16.4 v.s. 17.8). These experiments support that the X-DETR can serve as an effective open-vocabulary object detector.

\begin{table}[t]
\setlength{\tabcolsep}{2.5pt}
\begin{center}
\small
\begin{tabular}{lccccccc} 
\toprule
\multirow{2}{*}{Method} &\multirow{2}{*}{FT} 
&\multicolumn{2}{c}{COCO} 
&\multicolumn{2}{c}{Objects365}
&\multicolumn{2}{c}{OpenImages}  \\ [0.5ex] 
&  &AP &AP50 &AP &AP50 &AP &AP50 \\
\midrule
Faster R-CNN \cite{ren2015faster} &\cmark &38.1 &58.9 &19.7 &30.6 &20.6 &30.8\\\midrule
R-CLIP &\xmark &22.8 &34.6 &5.9 &9.2 &15.2 &23.1\\
R-CLIP+ &\xmark &24.4 &36.8 &6.5 &10.1 &16.0 &24.1\\
MDETR \cite{kamath2021mdetr} &\xmark &3.0 &3.9 &0.5 &0.7 &0.4 &0.5\\
X-DETR (ours) &\xmark  &26.5 &38.9 &5.7 &8.6 &4.8 &6.7\\
 \bottomrule
\end{tabular}
\caption{OVOD results on other datasets.  ``FT'' means being finetuned on the target dataset.}
\label{tab:ovod}%\vspace{-5mm}
\end{center}\vspace{-3mm}
\end{table}

To evaluate the transferring ability, Mask R-CNN \cite{he2017mask} trained on full data, using repeat factor sampling (RFS) for class imbalance, is regarded as a strong baseline, and a vanilla DETR pretrained on COCO as a transferring baseline. When finetuning, X-DETR still has very strong improvements over MDETR, 6.1/5.3/11.5 points for 1\%/10\%/100\% data. Using more data leads to worse results in MDETR (100\% v.s. 10\% data), showing that MDETR does not leverage detection data very well. But X-DETR has increasing gains with more detection data, whose result using 100\% data is better than the strong Mask R-CNN baseline (34.0 v.s. 33.3). Note that the X-DETR finetuning is straightforward and uses no strategy for the category imbalance issue.

In addition to OVID results on LVIS, we test X-DETR on COCO (80 classes) \cite{DBLP:conf/eccv/LinMBHPRDZ14}, OpenImages-v6 (601 classes) \cite{krasin2017openimages} and Objects365-v1 (365 classes) \cite{shao2019objects365}, in Table \ref{tab:ovod}. Although our pretraining uses the images from COCO and OpenImages, no category information is used. X-DETR achieves relatively good results on COCO (26.5 AP), but COCO only has 80 classes, and it is possible that the pretraining data has covered those categories information in the free-form language descriptions. Since Ojbects365 and OpenImages are much more challenging, the numbers of X-DETR are relatively lower, but are still much better than MDETR. R-CLIP has close results as X-DETR on COCO and Objects365, but much better on OpenImages ($\sim$10 points gain). These have shown that OVOD is still a very challenging task and requires more research efforts.

\begin{table*}[t]
\setlength{\tabcolsep}{4pt}
\begin{center}
\small
\begin{tabular}{lcc|ccc|ccc|ccc} 
\toprule
\multirow{2}{*}{Method} &\multirow{2}{*}{FT} &\multirow{2}{*}{Time} 
&\multicolumn{3}{c|}{RefCOCO val} 
&\multicolumn{3}{c|}{RefCOCO+ val}
&\multicolumn{3}{c}{RefCOCOg val}  \\ [0.5ex] 
& &  &R@5 &R@10 &R@30 &R@5 &R@10 &R@30 &R@5 &R@10 &R@30  \\
\hline
R-CLIP &\xmark &$\sim$0.19ms &5.6 &8.1  &14.8  &7.3  &10.2 &17.3 &21.7 &29.4  &42.9 \\
R-CLIP+ &\xmark &$\sim$0.19ms &5.0 &7.1  &12.8  &6.3  &9.0 &14.7 &20.0 &27.3  &40.6 \\
12-in-1 \cite{lu202012} &\xmark &$\sim$3.5s &1.0 &2.1  &5.8 &0.9  &1.8 &5.4 &2.7 &5.4  &12.9 \\
MDETR \cite{kamath2021mdetr} &\xmark &$\sim$25s &1.3 &2.5  &6.6  &1.1  &2.2 &5.4 &1.5 &2.8  &7.5 \\
X-DETR (ours) &\xmark &$\sim$0.15ms &21.5 &30.8  &47.8  &14.8  &22.1 &37.7 &23.4 &33.2  &52.0 \\\hline
UNITER \cite{DBLP:conf/eccv/ChenLYK0G0020} &\cmark &$\sim$1.4s  &8.1 &14.3  &28.9  &13.5  &21.0 &36.0 &14.5 &22.1  &37.7 \\
MDETR \cite{kamath2021mdetr} &\cmark &$\sim$25s  &2.0 &3.7  &9.0  &2.5  &4.4 &10.9 &3.5 &5.9  &15.3 \\
X-DETR (ours) &\cmark &$\sim$0.15ms  &29.9 &40.7  &59.6  &23.7  &33.5 &53.8 &40.0 &53.4  &72.5 \\
\bottomrule
\end{tabular}
\caption{Multi-modal instance search results. Time is evaluated per query on RefCOCO (1,500 images) after feature indexing.}
\label{tab:mmis}%\vspace{-3mm}
\end{center}\vspace{-3mm}
\end{table*}

%%%%%%%%%%%%%%%%%%%%%%%%% MMIS example %%%%%%%%%%%%%%%%%%%%%%%%%%%
\begin{figure*}[t]
\centering
\centerline{\includegraphics[width=1\linewidth]{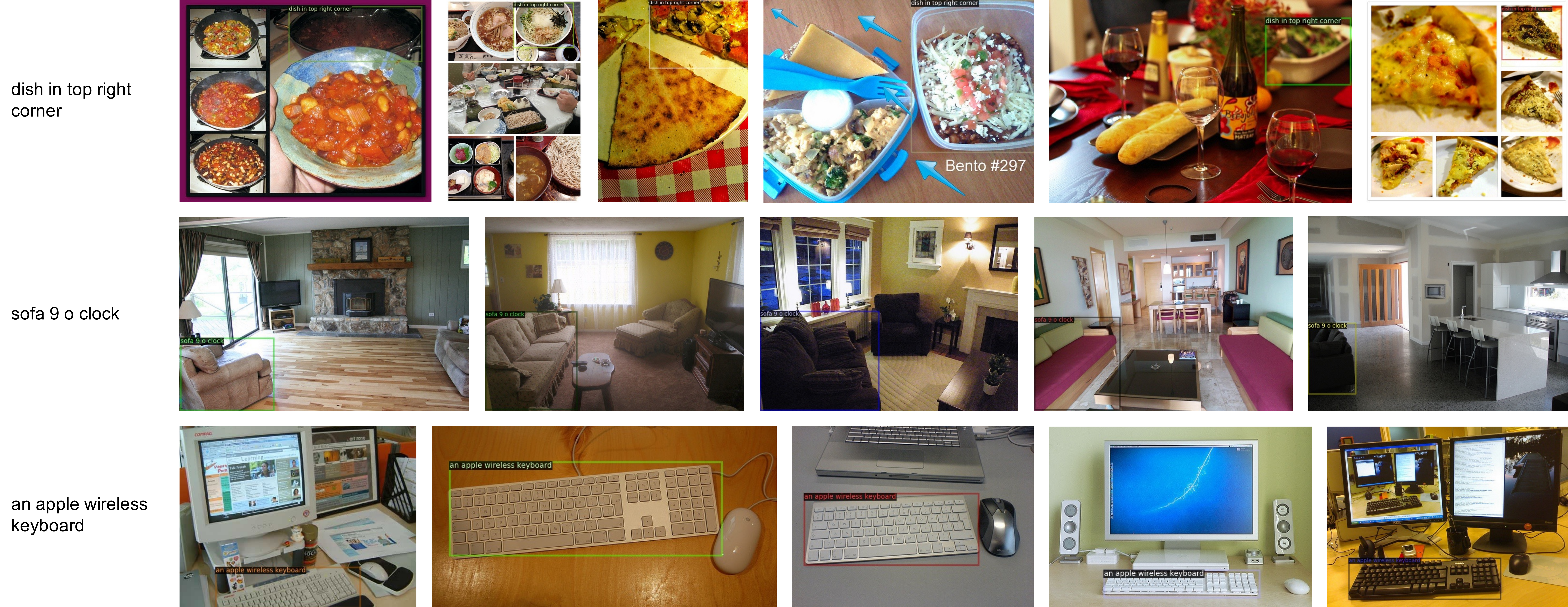}}
\caption{The top MMIS retrieval results for the queries from RefCOCO/RefCOCO+/RefCOCOg (from top to bottom), on COCO validation set.}
\label{fig:mmis examples}\vspace{-2mm}
\end{figure*}

%------------------------------------------------------------------------
\subsection{Multi-modal Instance Search (MMIS)}

Next, X-DETR is evaluated on MMIS \cite{hinami2017discriminative,plummer2020revisiting,liu2021ovis}. This 1) is a practical problem, like the commercial search engines (Google, Bing, etc.), 2) is challenging, as there are tons of false positives due to the  large-scale image database, and 3) requires high efficiency, i.e., instant retrieval results for millions/billions of images for commercial search engines. However, collecting MMIS datasets is very expensive, because it needs to annotate all bounding boxes in the full database for any free-form language query. This is a main reason that there is no publicly available dataset for this task yet\footnote{A recent work \cite{liu2021ovis} discusses this, but no data is released yet.}.

To evaluate on this task, we converted the referred expression comprehension (REC) datasets (RefCOCO/RefCOCO+/RefCOCOg \cite{kazemzadeh2014referitgame,DBLP:conf/eccv/YuPYBB16,DBLP:conf/cvpr/MaoHTCY016}), because REC is a special (simpler) case of MMIS, which is to search the result in a single image. We changed the evaluation protocol of REC, such that the retrieval operates on the full database instead of a single image. However, only one bounding box is associated with a given query in REC datasets, which is not exclusive across the full database. For example, multiple bounding boxes from different images could correspond to a single query, especially when the query is somewhat general, e.g., ``left man''.  Therefore, we used a loose evaluation metric, the recall @ the top \{5, 10, 30\} bounding boxes. The cross-modal similarity scores are used to rank and retrieve the objects for each query. The results shown in Table \ref{tab:mmis}, and some MMIS examples are shown in Figure \ref{fig:mmis examples}.

R-CLIP does not perform  well for RefCOCO/RefCOCO+, because these two datasets require more context information, e.g., ``left/right'', which is missing in R-CLIP. Even when including the context region, it does not improve. For RefCOCOg, which focuses more on general language description, which CLIP was mainly trained on, R-CLIP shows much better results. On the contrary, X-DETR is better in all three datasets, especially on RefCOCO/RefCOCO+.  When compared with 12-in-1 \cite{lu202012}, UNITER \cite{DBLP:conf/eccv/ChenLYK0G0020} and MDETR \cite{kamath2021mdetr}, which use  cross-modality transformers to model vision-language alignment and achieved very good results on REC tasks, X-DETR outperforms them by a large margin with and without finetuning.
These observations are consistent with \cite{plummer2020revisiting}: a good model for REC is not necessarily a  good model for the MMIS task.

\begin{table*}[t]
\setlength{\tabcolsep}{5pt}
\begin{center}
\small
\begin{tabular}{lccccccccccccc} 
\toprule
\multirow{2}{*}{Method} &\multirow{2}{*}{FT}
&\multicolumn{3}{c}{RefCOCO val} 
&\multicolumn{3}{c}{RefCOCO+ val}
&\multicolumn{3}{c}{RefCOCOg val}  \\ [0.5ex] 
&  &R@1 &R@5 &R@10 &R@1 &R@5 &R@10 &R@1 &R@5 &R@10  \\
\midrule
R-CLIP &\xmark & 21.6 & 52.2 & 68.5  & 24.7 & 57.4 & 72.8 & 36.4 & 72.3 & 86.4 \\
R-CLIP+ &\xmark & 17.3 & 45.6 & 62.0  & 19.8 & 48.6 & 65.4 & 32.4 & 68.0 &82.7 \\
MDETR \cite{kamath2021mdetr} &\xmark &72.4 &92.2 &94.7  &58.3  &86.3 &90.5  &55.9 &87.0 &91.8  \\
X-DETR (ours) &\xmark  &78.7 &95.4 &97.6  &63.5  &92.5 &96.2  &60.4 &91.8 &95.6 \\\hline
MAttNet\cite{yu2018mattnet} &\cmark  &76.7  & - & - & 65.3 & - & - & 66.6 & -  \\
UNITER$_L$\cite{DBLP:conf/eccv/ChenLYK0G0020}$^*$ &\cmark & 81.4 & -  & - & 75.9  & - & - & 74.9 & - \\  
VILLA$_L$\cite{gan2020large}$^*$ &\cmark & 82.4 & -  & - &  76.2 & - &  - & 76.2 & -   \\
MDETR \cite{kamath2021mdetr} &\cmark  & \textbf{86.8} &96.0 &97.2    & \textbf{79.5}  &96.2 &97.5  & \textbf{81.6} &95.5 &96.8  \\
X-DETR (ours) &\cmark  &86.2 &\textbf{97.8} &\textbf{98.9}  &77.0  &\textbf{97.1} &\textbf{98.6}  &80.4 &\textbf{96.8} &\textbf{97.9}  \\
 \bottomrule
\end{tabular}
\caption{Comparison with state-of-the-art on REC datasets.}
\label{tab:refexp}%\vspace{-3mm}
\end{center}\vspace{-6mm}
\end{table*}

\setlength{\tabcolsep}{5pt}
\begin{table}[t]
\begin{center}
\small
\begin{tabular}{lcccc}
\toprule
Method &FT & R@1 & R@5 & R@10\\\hline
R-CLIP &\xmark &21.9 &48.4 &60.0 \\
MDETR \cite{kamath2021mdetr} &\xmark & \textbf{82.5} & 92.9 &94.9\\
X-DETR (ours) &\xmark  &81.4 &\textbf{93.6} &\textbf{95.6}\\\hline
VisualBert\cite{li2019visualbert} &\cmark &  68.1 &  84.0 & 86.2 \\
VisualBert$\dagger$\cite{li2019visualbert}  &\cmark &  70.4 &  84.5 & 86.3 \\
X-DETR (ours) &\cmark  &81.8 &\textbf{93.6} &95.5 \\
\bottomrule
\end{tabular}
\caption{Phrase grounding results on Flickr30k entities validation.}
\label{tab:flickr}
\end{center}\vspace{-3mm}
\end{table}

%------------------------------------------------------------------------
\subsection{Phrase Grounding and Referring Expression Comprehension}

Phrase grounding and referring expression comprehension (REC) are  reduced versions of MMIS, which retrieve the targets most similar to the given query in a single image. They are simpler tasks than MMIS, which assumes that the object referred by the query definitely exists in the image.

Flickr30k entities dataset \cite{DBLP:journals/ijcv/PlummerWCCHL17} is used for phrase grounding evaluation, with the train/val/test splits of \cite{DBLP:journals/ijcv/PlummerWCCHL17} and Recall@\{1, 5, 10\} as the evaluation metrics. For finetuning, the pretrained X-DETR model was finetuned for 3 epochs on the target dataset, with the same losses as pretraining. The results are shown in Table \ref{tab:flickr}. R-CLIP performs poorly on this task. Compared with MDETR \cite{kamath2021mdetr}, X-DETR only has a small gap for R@1 but better performance at R@5 and R@10, considering X-DETR does not use the much stronger transformer as the joint-modality modeling. Finetuning also helps X-DETR, but does not help MDETR as mentioned in \cite{kamath2021mdetr}.

RefCOCO, RefCOCO+ and RefCOCOg \cite{kazemzadeh2014referitgame,DBLP:conf/eccv/YuPYBB16,DBLP:conf/cvpr/MaoHTCY016} are used for REC evaluation with Recall@\{1, 5, 10\} as the evaluation metrics. 
For finetuning, the pretrained model was finetuned for 4 epochs on the union of all three dataset (excluding all images in all three validation sets), with the object detection and object-sentence alignment losses introduced in Section \ref{sec:training}. The results are shown in Table \ref{tab:refexp}. Similar to grounding in Table \ref{tab:flickr}, R-CLIP does not work very well for REC task. X-DETR outperforms UNITER$_L$ \cite{DBLP:conf/eccv/ChenLYK0G0020} and VILLA$_L$ \cite{gan2020large}, on all three datasets. When compared with the current state-of-the-art MDETR \cite{kamath2021mdetr} on the finetuning setting, X-DETR is slightly worse at R@1 but better at R@5 and R@10. When the model is not finetuned, X-DETR achieves much better results than MDETR. These results have shown that X-DETR can be generalized well for simpler tasks such as phrase grounding and REC, in addition to the challenging tasks of OVOD and MMIS.

%------------------------------------------------------------------------
\subsection{Speed Comparisons}

X-DETR is an efficient architecture for both training and inference. When compared with R-CLIP, X-DETR is about 100 times faster for OVOD due to the slow R-CNN pipeline. For MMIS, X-DETR has close retrieval time with R-CLIP, since they both use the simple dot-product as the vision-language alignment and vision/language features are fully indexable. However, the indexing speed of X-DETR is about 100 times faster than R-CLIP due to the R-CNN pipeline. When compared with the state-of-the-art localization based V+L work of MDETR, X-DETR is a few times faster during training, since it can process an image with all of its queries simultaneously but MDETR needs to process the queries one by one due to the joint-modality transformer design, as discussed in Section \ref{subsec:training efficiency}. For example, finetuning on 10\% LVIS, MDETR needs 108 hours for 150 epochs, but X-DETR only needs 5.2 hours for 50 epochs with improved results (see Table \ref{tab:lvis}). At inference, for OVOD/MMIS, X-DETR is about 100/100,000 times faster than MDETR (see Table \ref{tab:lvis} and \ref{tab:mmis}). Note that we have already indexed the indexable features, e.g., the ones before the joint-modality transformers for MDETR. The training (inference) speeds are reported on 8 (1) A100 GPUs.

\begin{table}[t]
\setlength{\tabcolsep}{3pt}
\begin{center}
\small
\begin{tabular}{lccccccccccc} 
\toprule
Pre-training &\multirow{2}{*}{FT} 
&\multicolumn{2}{c}{LVIS}
&\multicolumn{2}{c}{Refg*} &Flickr &Ref
\\ [0.5ex] 
data & & AP & AP@r & R@5 & R@10 & R@1 & R@1 \\
\midrule
\emph{mixed} &\xmark  &11.5  &6.0  &16.2 &24.6 &79.7 &63.9 \\
\emph{mixed}* &\xmark  &13.5  &5.2  &20.4 &29.8 &79.9 &75.4 \\
\emph{mixed}*+boxes &\xmark &14.7  &5.1  &22.4 &32.4 &79.9 &78.0 \\
+CC &\xmark  &15.9  &7.9  &22.2 &31.8 &80.1 &76.9 \\
+CC+LocNar &\xmark  &15.7  &6.8  &21.0 &29.8 &81.3 &78.3 \\
\midrule
\emph{mixed} &\cmark  &20.0 &13.4 &34.3 &45.8 &80.1 &83.0 \\
\emph{mixed}* &\cmark  &21.7 &18.6 &36.2 &48.9 &80.6 &83.9 \\
\emph{mixed}*+boxes &\cmark  &22.8 &14.8 &36.5 &49.4 &80.3 &84.1 \\
+CC &\cmark  &23.3 &16.6 &37.5 &49.4 &80.9 &84.5 \\
+CC+LocNar &\cmark  &25.6 &17.3 &37.5 &50.3 &81.8 &85.4 \\
\bottomrule
\end{tabular}
\caption{Ablation studies. ``Ref'' means RefCOCO, ``Refg*'' means MMIS on RefCOCOg. Finetuning on LVIS is on 10\% data.}
\label{tab:ablation}%\vspace{-5mm}
\end{center}\vspace{-3mm}
\end{table}

%------------------------------------------------------------------------
\subsection{Ablation on Pretraining Data}

We ablated the effect of the pretraining data in Table \ref{tab:ablation} (with ResNet-50 for efficiency purpose). The rows are for experiments using data of 1) the original \emph{mixed} of MDETR \cite{kamath2021mdetr} (``\emph{mixed}''), where all queries of an image are merged into a single paragraph, violating the independence assumption of the queries; 2) \emph{mixed} with independent queries (``\emph{mixed}*''), by splitting the merged paragraph into independent queries; 3) adding COCO objects without category (``\emph{mixed}*+boxes''); 4) adding CC  \cite{DBLP:conf/acl/SoricutDSG18} weakly labeled image-caption data (``+CC''); 5) adding LocNar \cite{DBLP:conf/eccv/Pont-TusetUCSF20} pseudo-labeled and OpenImage \cite{krasin2017openimages} bounding box data (``+CC+LocNar''). It can be found that the original \emph{mixed} dataset has a mismatch with the inference data, which leads to inferior results especially when the model is not finetuned. Independent queries of \emph{mixed*} can have significant accuracy boosts in multiple tasks. Adding COCO object data (``\emph{mixed}*+boxes'') also has nontrivial improvements, showing the advantage of X-DETR to leverage detection-only data. When using CC data (``+CC''), the results of OVOD on LVIS are improved because more concepts are covered by more data, but the results on RefCOCO are decreased. When adding more LocNar pseudo-labeled data with OpenImage objects (``+CC+LocNar''), the results on Flickr30k and RefCOCO are improved over ``+CC'', but decreased on MMIS. The observations on finetuning settings are consistent. These ablation studies have shown that it is not enough to just use the fully annotated object-phrase annotations, and leveraging other weaker types of supervision could be helpful for instance-wise V+L learning.

%=======================================================================
\section{Conclusion}

In this paper, we propose a simple yet effective architecture for instance-wise vision-language tasks, which uses dot-product to align vision and language. It has shown that the expensive joint-modality transformer may not be necessary for those V+L tasks and the weak annotated data can be a big help to improve the model performances. The gains of the proposed X-DETR in terms of accuracy and speed are significant, when compared to the previous V+L language state-of-the-art, on the practical and challenging tasks such as open-vocabulary detection and multi-modal instance search. 

%\clearpage
%\newpage
\appendix

%%%%%%%%%%%%%%%%%%%%%%%%% data examples %%%%%%%%%%%%%%%%%%%%%%%%%%%
%\vspace{-10mm}
\begin{figure*}[t!]
\begin{minipage}[b]{.32\linewidth}
\centering
\centerline{\includegraphics[width=1.0\linewidth]{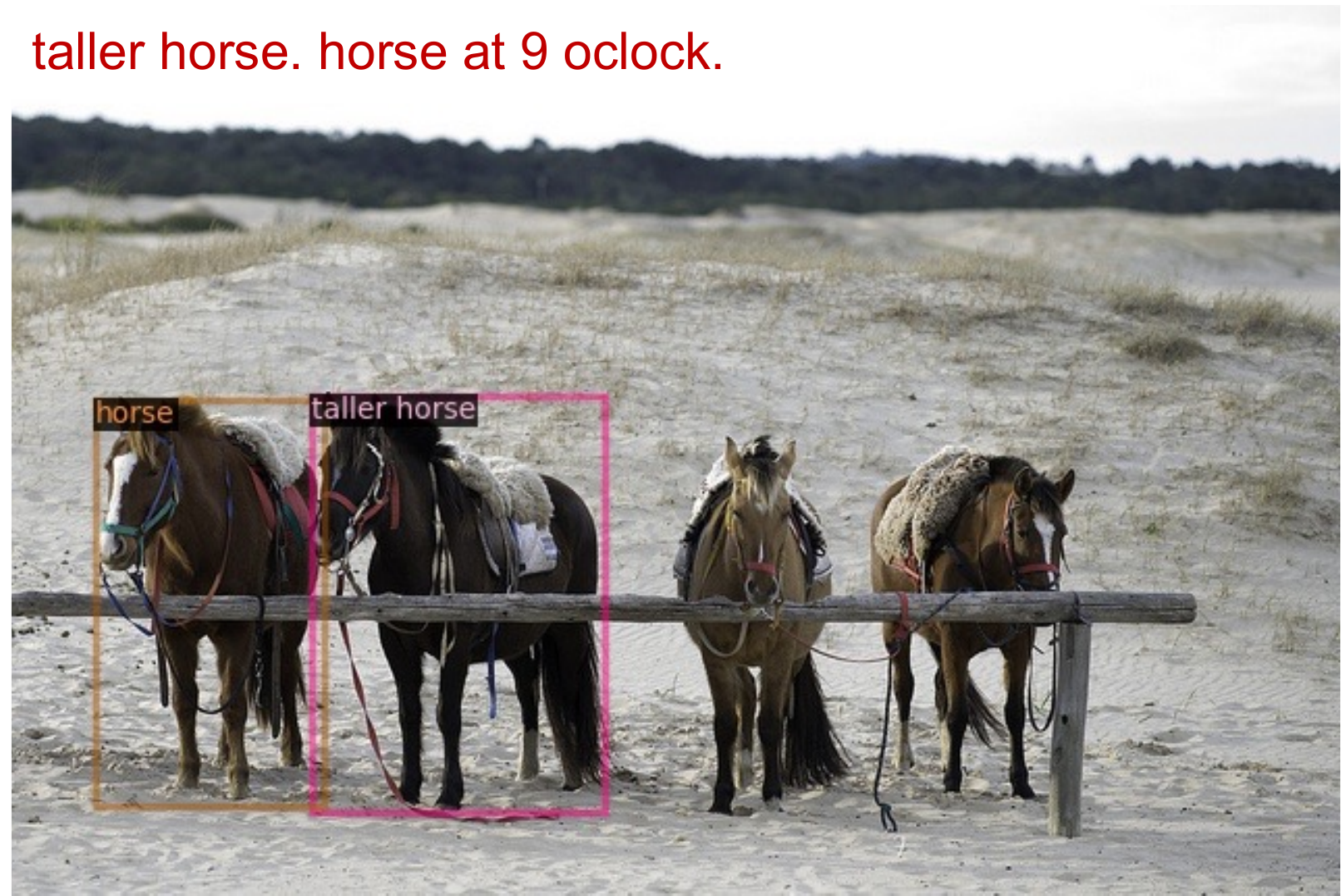}}{(a) Example of \emph{mixed}}
\end{minipage}
\hfill
\begin{minipage}[b]{.32\linewidth}
\centering
\centerline{\includegraphics[width=1.0\linewidth]{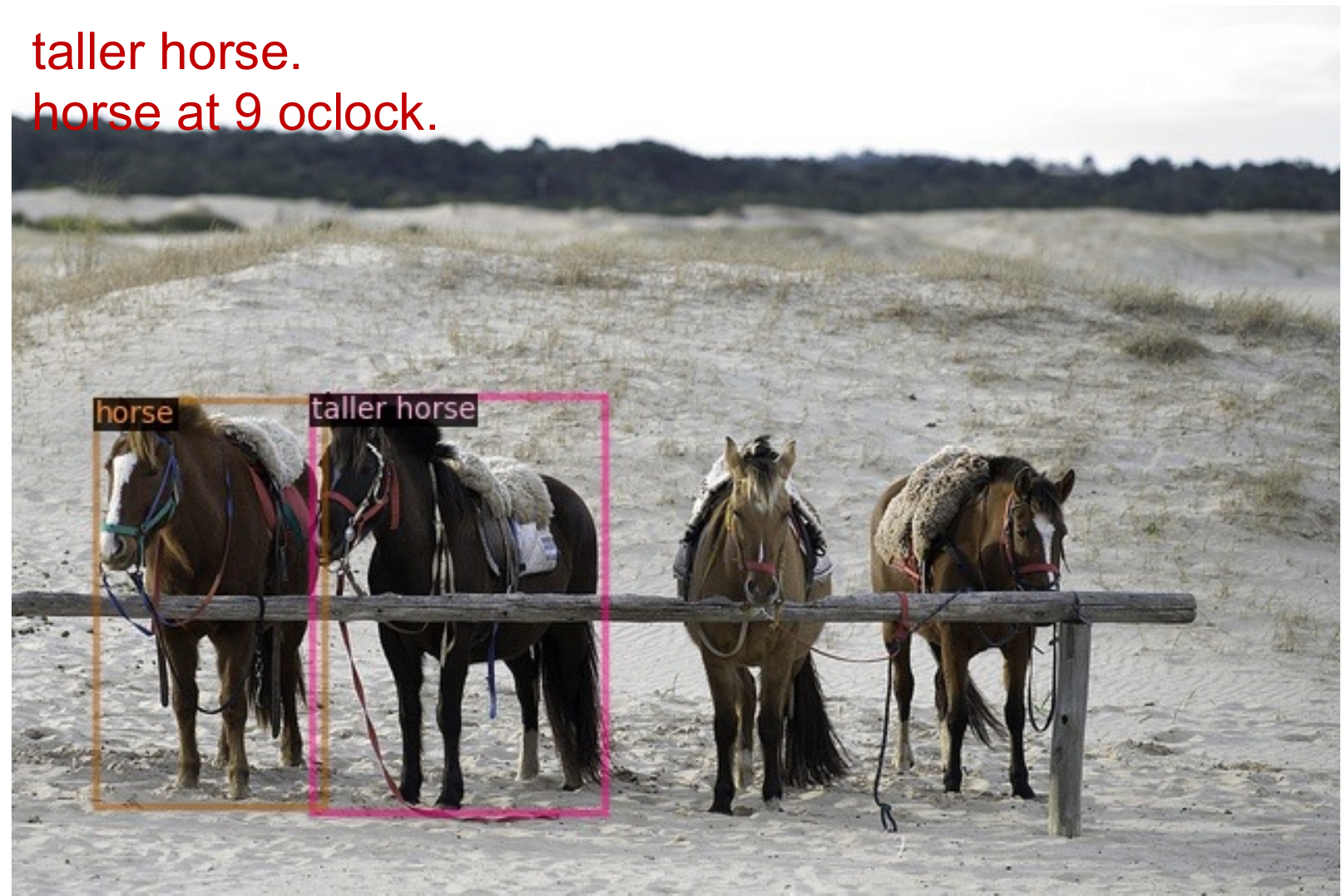}}{(b) Example of \emph{mixed}*}
\end{minipage}
\hfill
\begin{minipage}[b]{.32\linewidth}
\centering
\centerline{\includegraphics[width=1.0\linewidth]{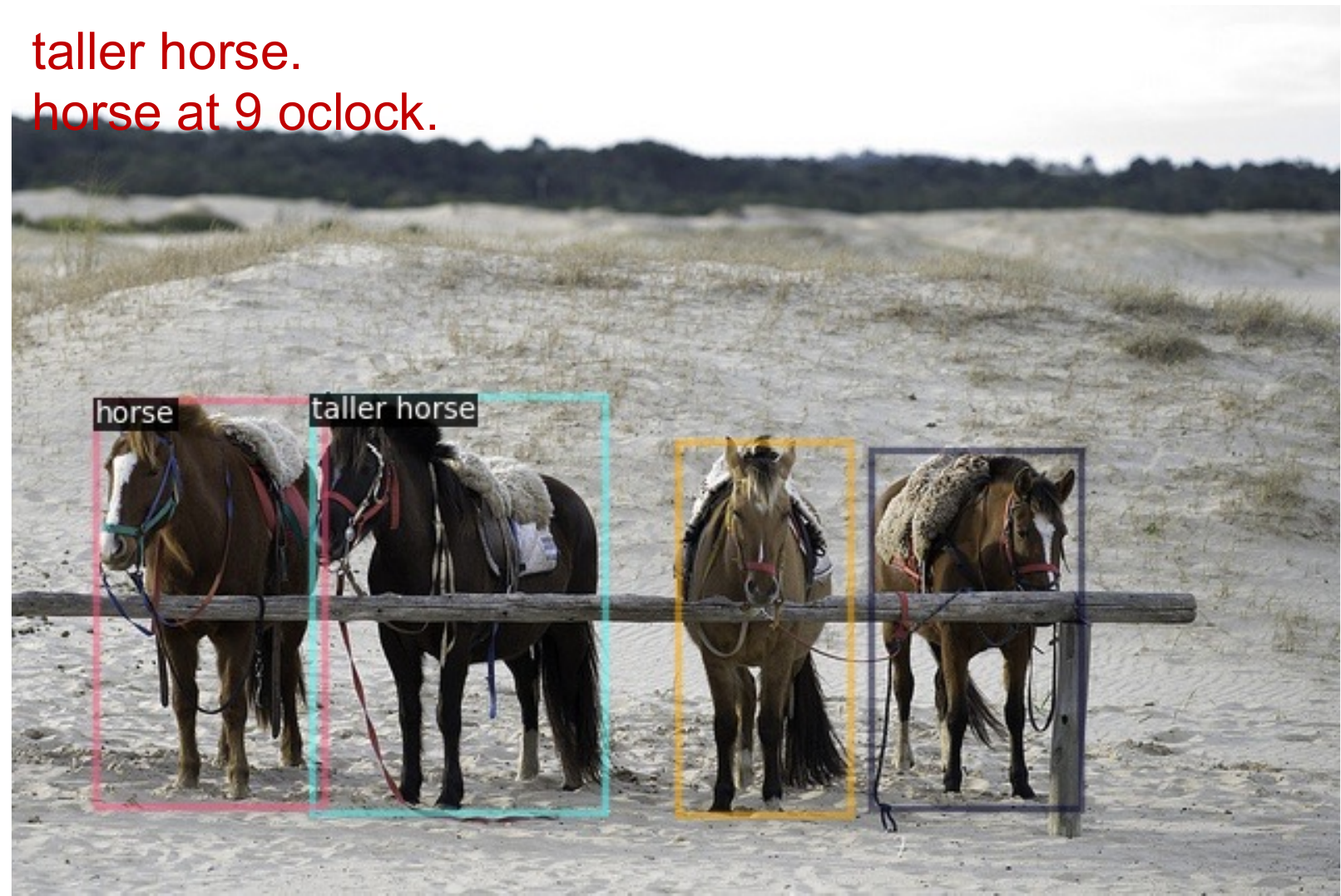}}{(c) Example of \emph{mixed}*+boxes}
\end{minipage}
\caption{An image example for \emph{mixed} (a), \emph{mixed}* (b) and \emph{mixed}*+boxes (c) of Table \ref{tab:ablation}. The text query is shown on the top left corner of each image. The text description on each bounding box is the noun phrase extracted from the text query. In the original \emph{mixed} dataset of MDETR, the independent queries are merged into a single query (one sentence) and objects are sparsely annotated (annotations for the right two horses are missing), as in (a). We at first split the paragraph query into independent queries (two sentences) as in (b). Then we add COCO bounding boxes (right two horses) to the dataset withouht category information (no text description on the added bounding boxes), as in (c).
}
\label{fig:examples}
\end{figure*}

%%%%%%%%% BODY TEXT
%=======================================================================
\section*{\Large{Appendix}}
In the appendix, we provide more details and visualization examples which are not covered in the main paper.

\section{Training Details}

\subsection{Pretraining}

During pre-training on the joint datasets, X-DETR was trained for 10 epochs w.r.t. the \emph{mixed} dataset. The batch size for fully/pseudo/weakly-annotated data is 4/2/4 for a single GPU, and we used 8 GPUs for training. The initial base learning rate is $1\times{10}^{-5}$ for backbone, $2.5\times{10}^{-5}$ for text encoder, $1\times{10}^{-5}$ for linear projection layers of Deformable DETR, and $1\times{10}^{-4}$ for the rest of parameters. And we followed the linear learning rate scaling rule: $lr=base\_lr\times{batch\_size}/16$, where $batch\_size$ is the batch size of fully-annotated data. The text encoder uses linear learning rate decay with warmup schedule, and the rest uses step learning rate decay, with learning rate dropped after the 8th epoch. All parameters are optimized by Adam with weight decay of $1\times{10}^{-4}$. The inference model is exponential moving averaged (EMA) from the model trajectory during training with
a decay rate of 0.9998.

\subsection{Finetuning on LVIS}

When finetuning on LVIS, X-DETR was trained for 50 epochs for 1\%/10\%/100\% data with batch size of 4 on each GPU. The initial base learning rate is $1\times{10}^{-5}$ for backbone, $1\times{10}^{-5}$ for text encoder, $5\times{10}^{-6}$ for linear projection layers of Deformable DETR, and $5\times{10}^{-5}$ for the rest of parameters. The learning rate dropped after the 40th epoch for step learning rate schedule. The image is resized such that the
minimum of width and height is 800. The other settings are the same as pretraining. We used the category names as the language description of the object, but remove the text in the parentheses, e.g., ``flip-flop\_(sandal)'' to ``flip-flop''.

\subsection{Finetuning on Flickr30k}

When finetuning on Flickr30k, X-DETR was trained for 3 epochs, with batch size of 4 on each GPU. The initial base learning rate is $5\times{10}^{-6}$ for backbone, $5\times{10}^{-6}$ for text encoder, $2.5\times{10}^{-6}$ for linear projection layers of Deformable DETR, and $2.5\times{10}^{-5}$ for the rest of parameters. The learning rate dropped after the 2nd epoch for step learning rate schedule. The other settings are the same as pretraining.

\subsection{Finetuning on REC Datasets}

When finetuning on REC datasets, we merged the RefCOCO/RefCOCO+/RefCOCOg together, excluding all images in all three validation sets. X-DETR was trained for 4 epochs, with batch size of 4 on each GPU. The initial base learning rate is $1\times{10}^{-6}$ for for backbone, $1\times{10}^{-5}$ for text encoder, $5\times{10}^{-6}$ for linear projection layers of Deformable DETR, and $5\times{10}^{-5}$ for the rest of parameters. The learning rate dropped after the 3rd epoch for step learning rate schedule. The other settings are the same as pretraining.

%%%%%%%%%%%%%%%%%%%%%%%%% LocNar examples %%%%%%%%%%%%%%%%%%%%%%%%%%%
\begin{figure*}[t]
\begin{minipage}[b]{.212\linewidth}
\centering
\centerline{\includegraphics[width=1.0\linewidth]{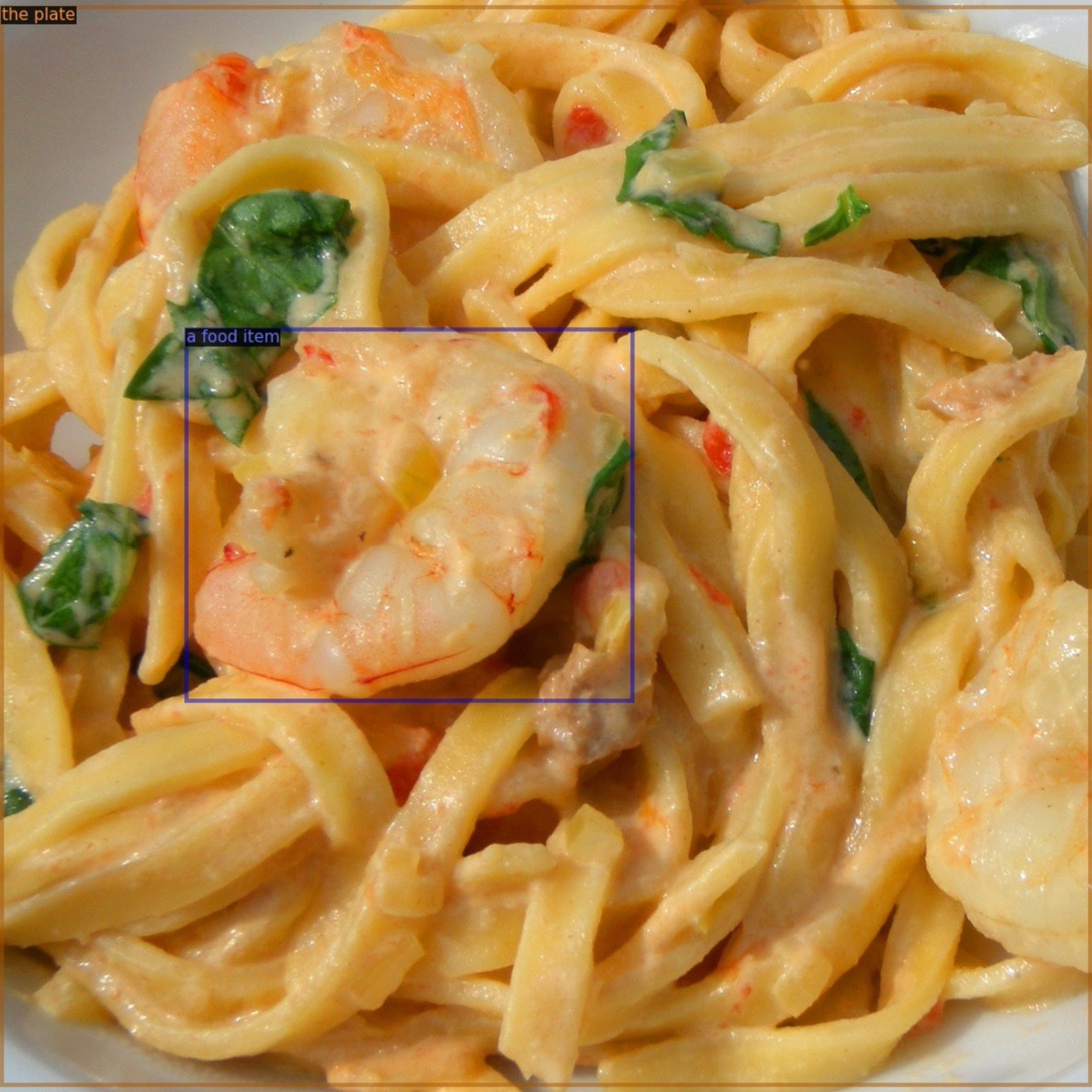}}
\end{minipage}
\hfill
\begin{minipage}[b]{.283\linewidth}
\centering
\centerline{\includegraphics[width=1.0\linewidth]{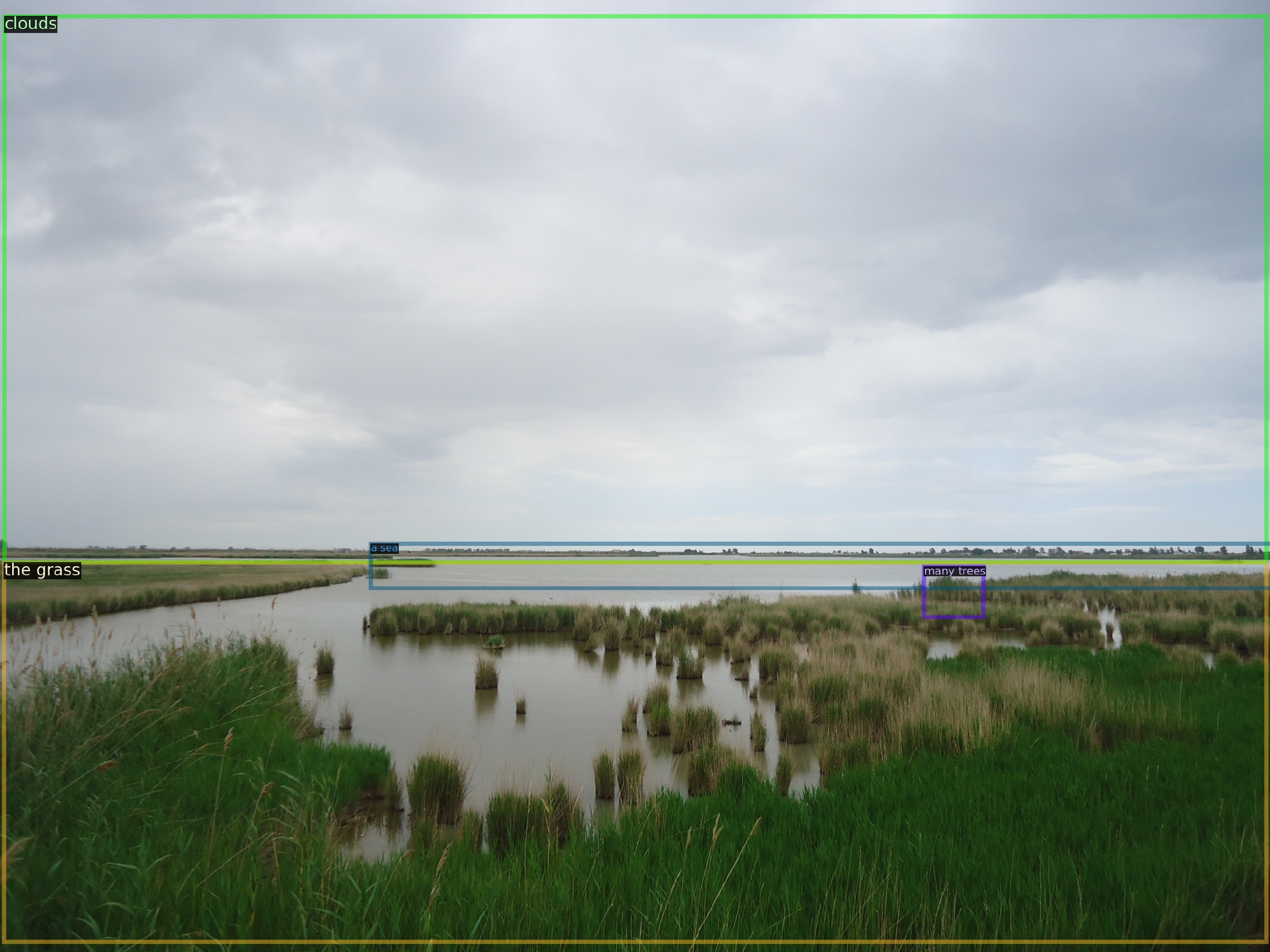}}
\end{minipage}
\hfill
\begin{minipage}[b]{.283\linewidth}
\centering
\centerline{\includegraphics[width=1.0\linewidth]{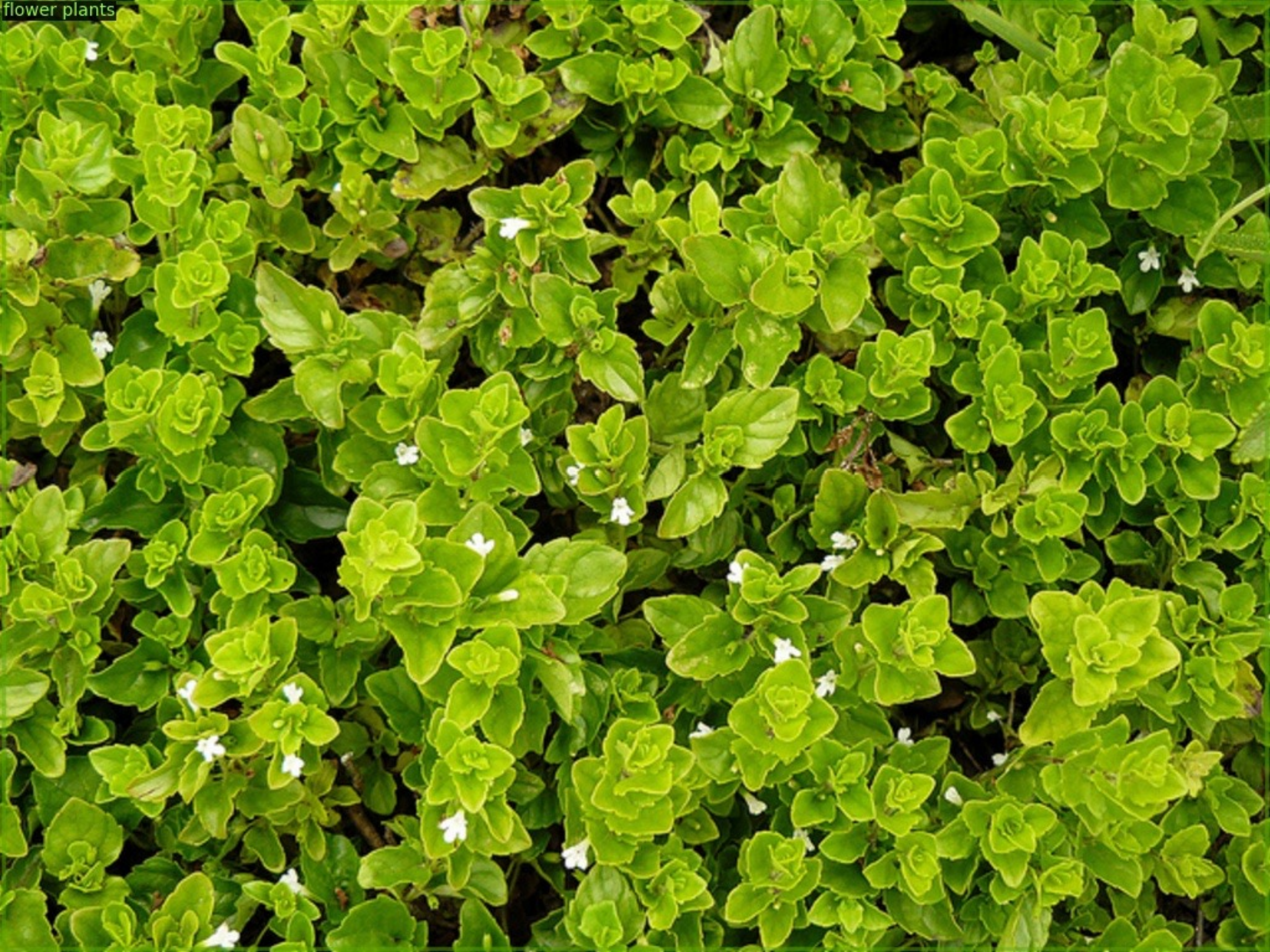}}
\end{minipage}
\hfill
\begin{minipage}[b]{.201\linewidth}
\centering
\centerline{\includegraphics[width=1.0\linewidth]{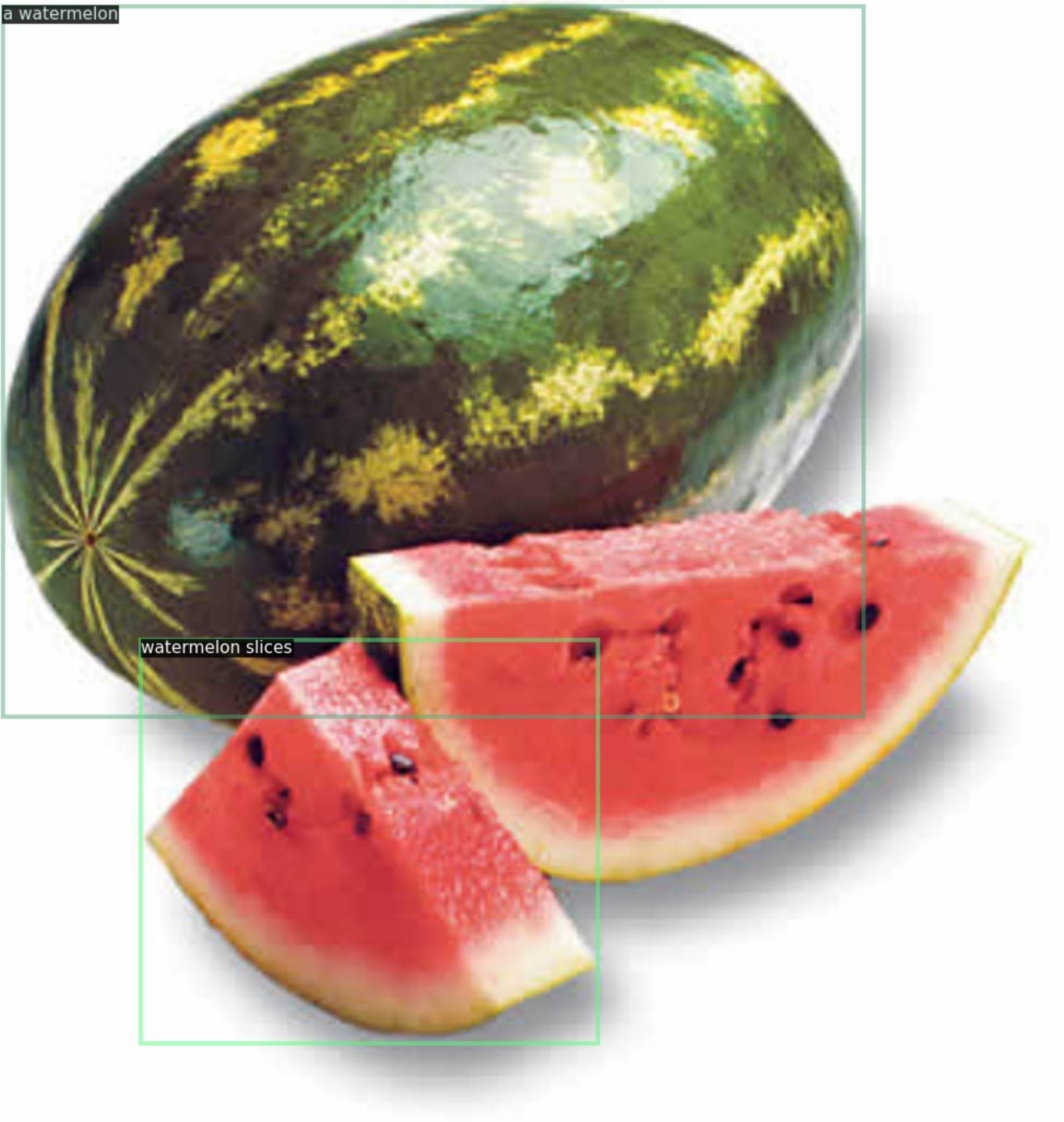}}
\end{minipage}\\
\begin{minipage}[b]{.212\linewidth}
\centering
\centerline{\includegraphics[width=1.0\linewidth]{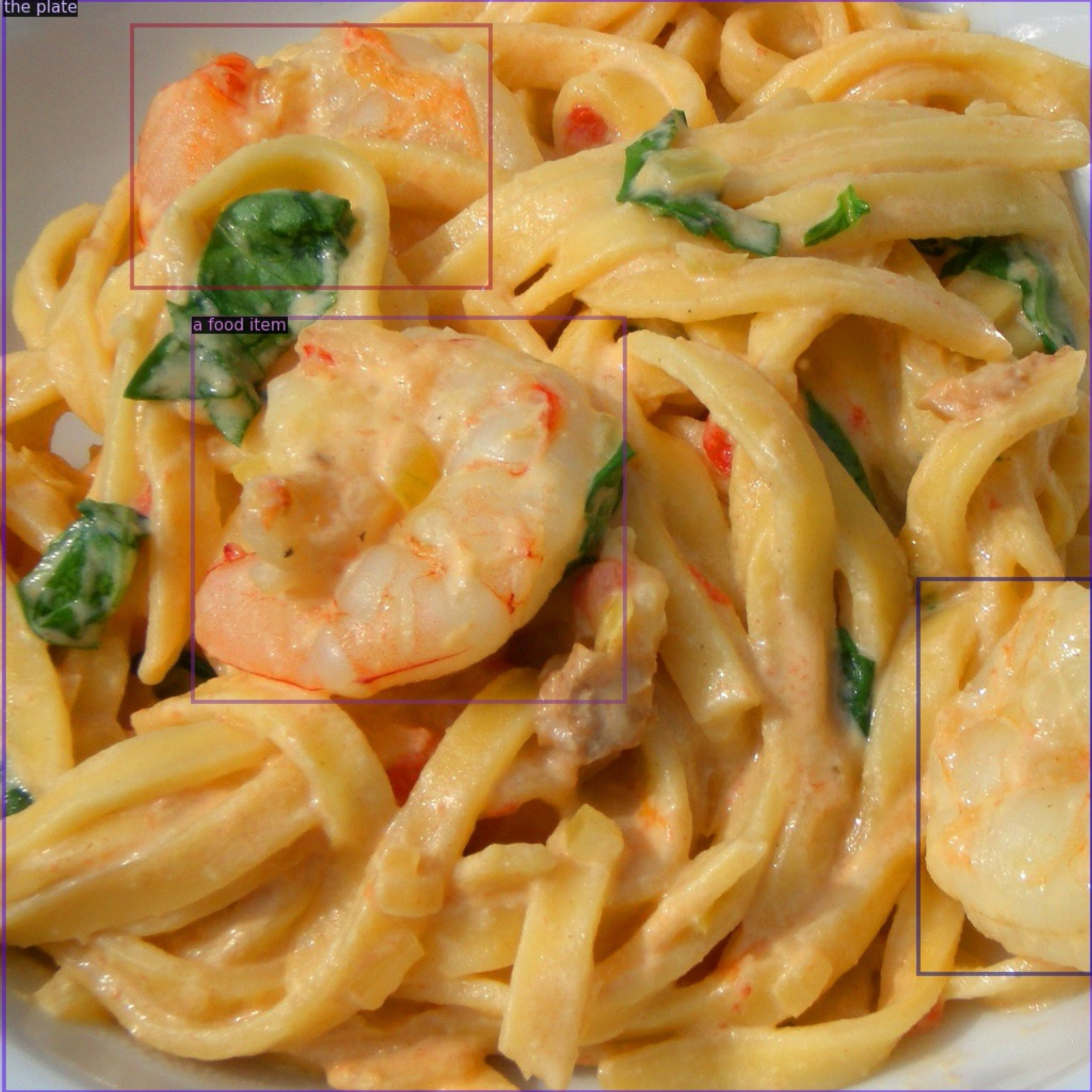}}{(a)}
\end{minipage}
\hfill
\begin{minipage}[b]{.283\linewidth}
\centering
\centerline{\includegraphics[width=1.0\linewidth]{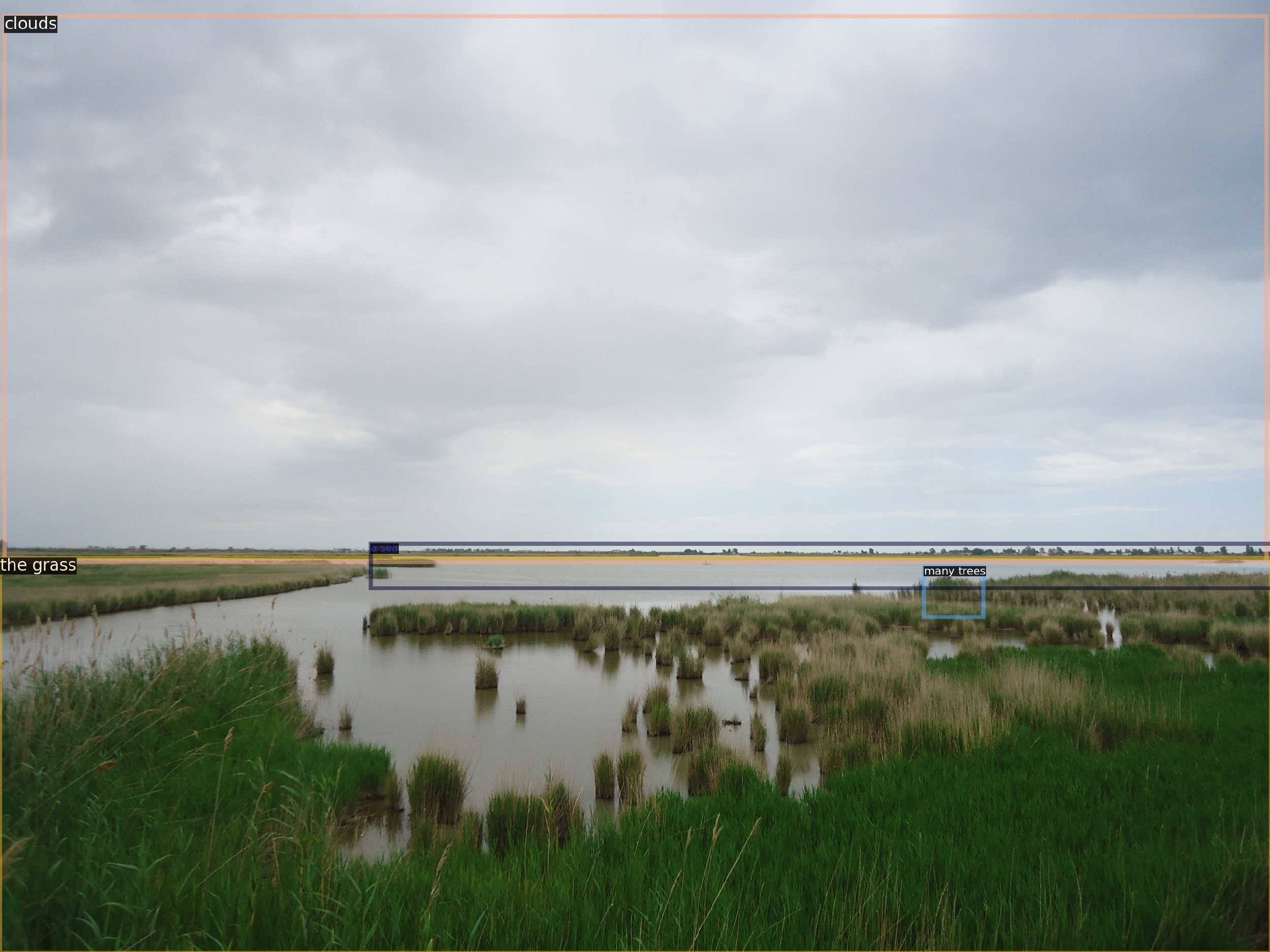}}{(b)}
\end{minipage}
\hfill
\begin{minipage}[b]{.283\linewidth}
\centering
\centerline{\includegraphics[width=1.0\linewidth]{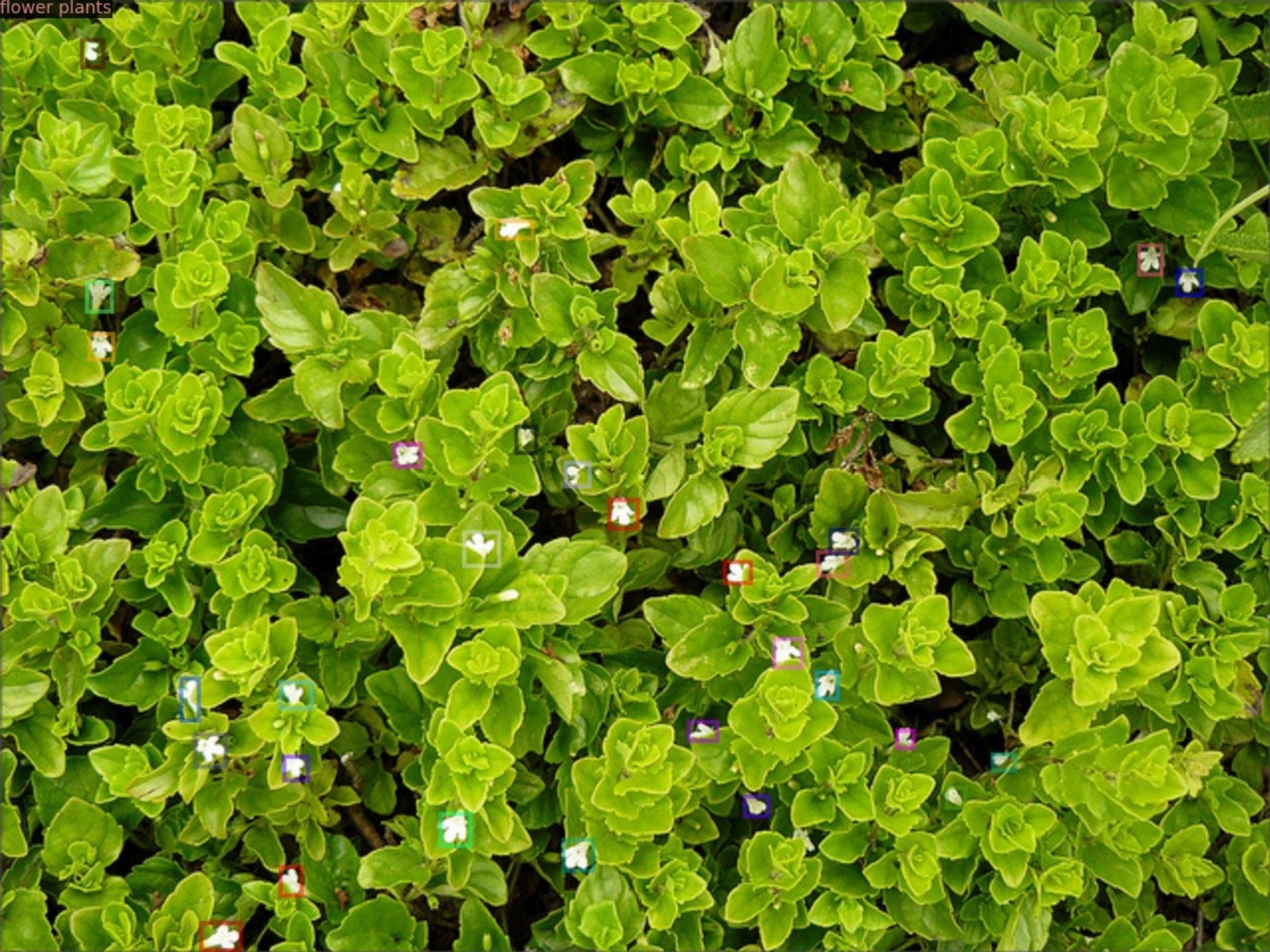}}{(c)}
\end{minipage}
\hfill
\begin{minipage}[b]{.201\linewidth}
\centering
\centerline{\includegraphics[width=1.0\linewidth]{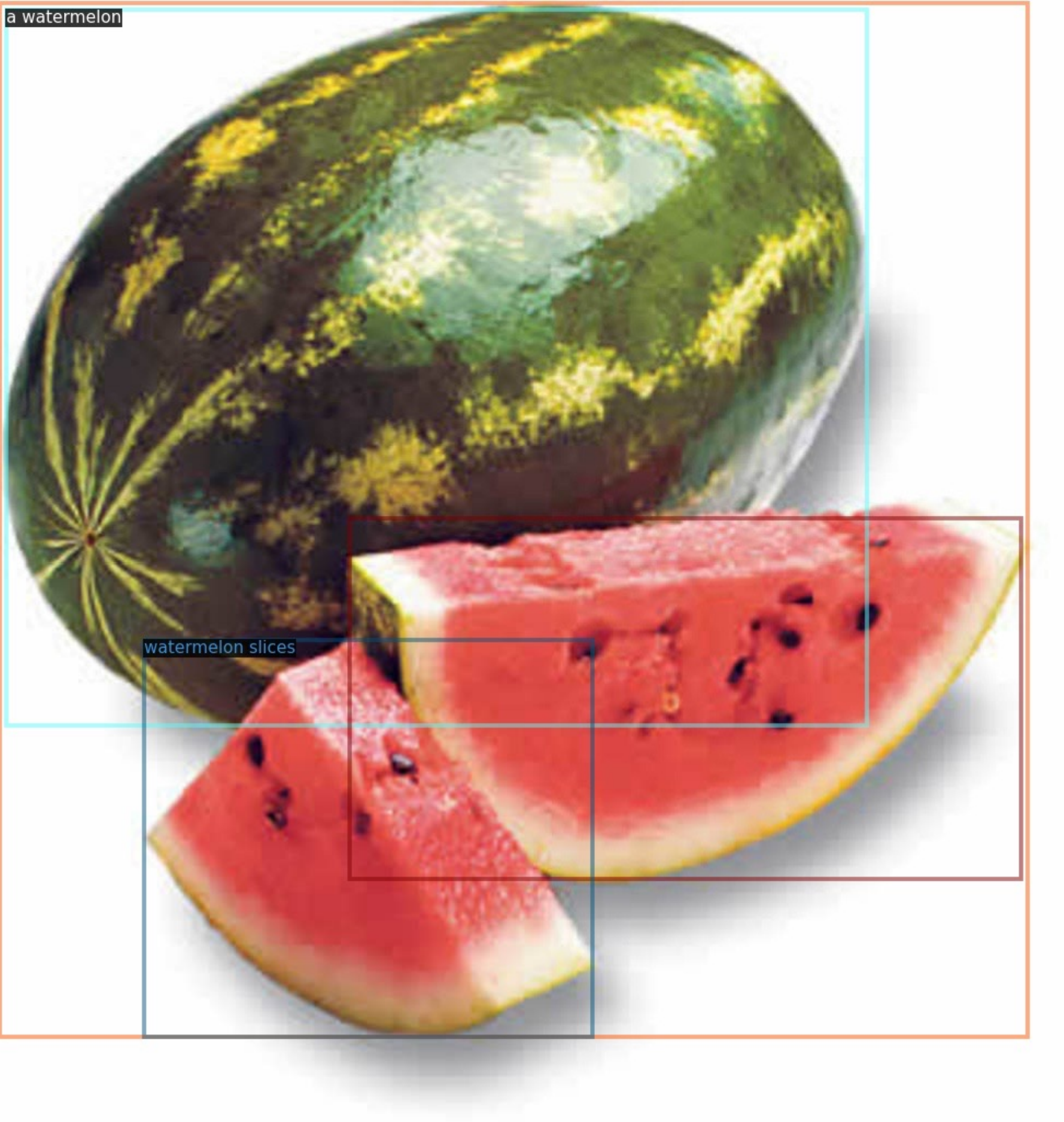}}{(d)}
\end{minipage}
\caption{The top images are pseudo label examples of LocNar, and the bottom images are the examples after adding OpenImage bounding boxes without category information (no text description on the added bounding boxes). The query lists are as follows,\\(a): [`there is a food item on the plate.']\\(b): [`This is an outside view.', `At the bottom, the grass.', `In the middle of the image there is a sea.', `In the background there are many trees.', `At the top of the image the sky and clouds.']\\(c): [`flower plants.']\\(d): [`a watermelon and watermelon slices.']
}
\label{fig:locnar}\vspace{-3mm}
\end{figure*}

%%%%%%%%%%%%%%%%%%%%%%%%% MMIS examples %%%%%%%%%%%%%%%%%%%%%%%%%%%
\begin{figure*}[t]
\begin{minipage}[b]{1.\linewidth}
\centering
\centerline{\includegraphics[width=1.0\linewidth]{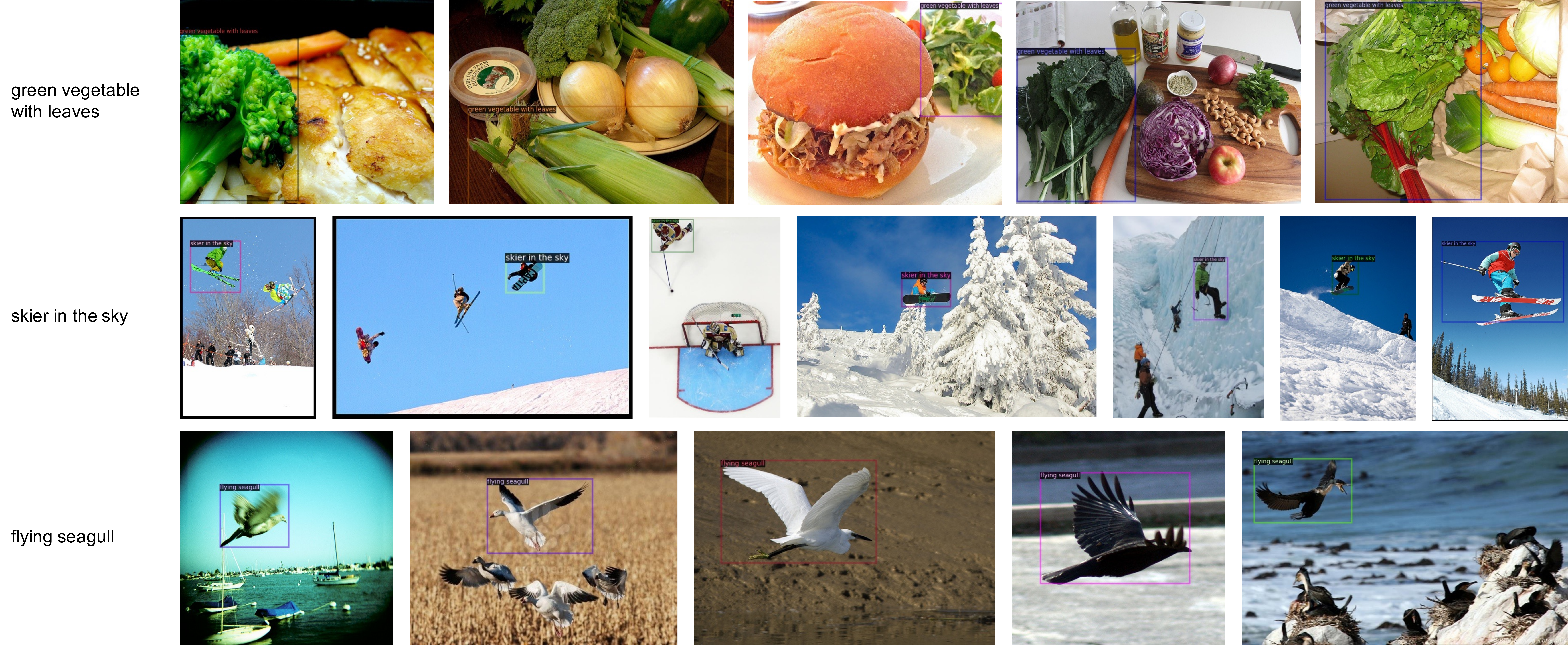}}\vspace{1mm}
\end{minipage}\\
\begin{minipage}[b]{1.\linewidth}
\centering
\centerline{\includegraphics[width=1.0\linewidth]{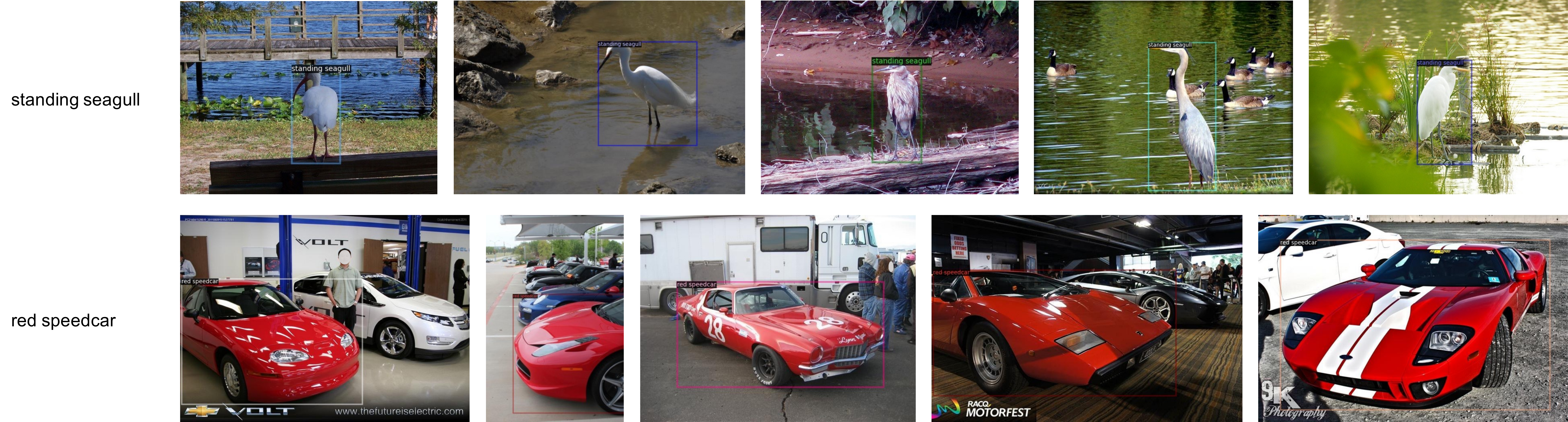}}\vspace{1mm}
\end{minipage}\\
\begin{minipage}[b]{1.\linewidth}
\centering
\centerline{\includegraphics[width=1.0\linewidth]{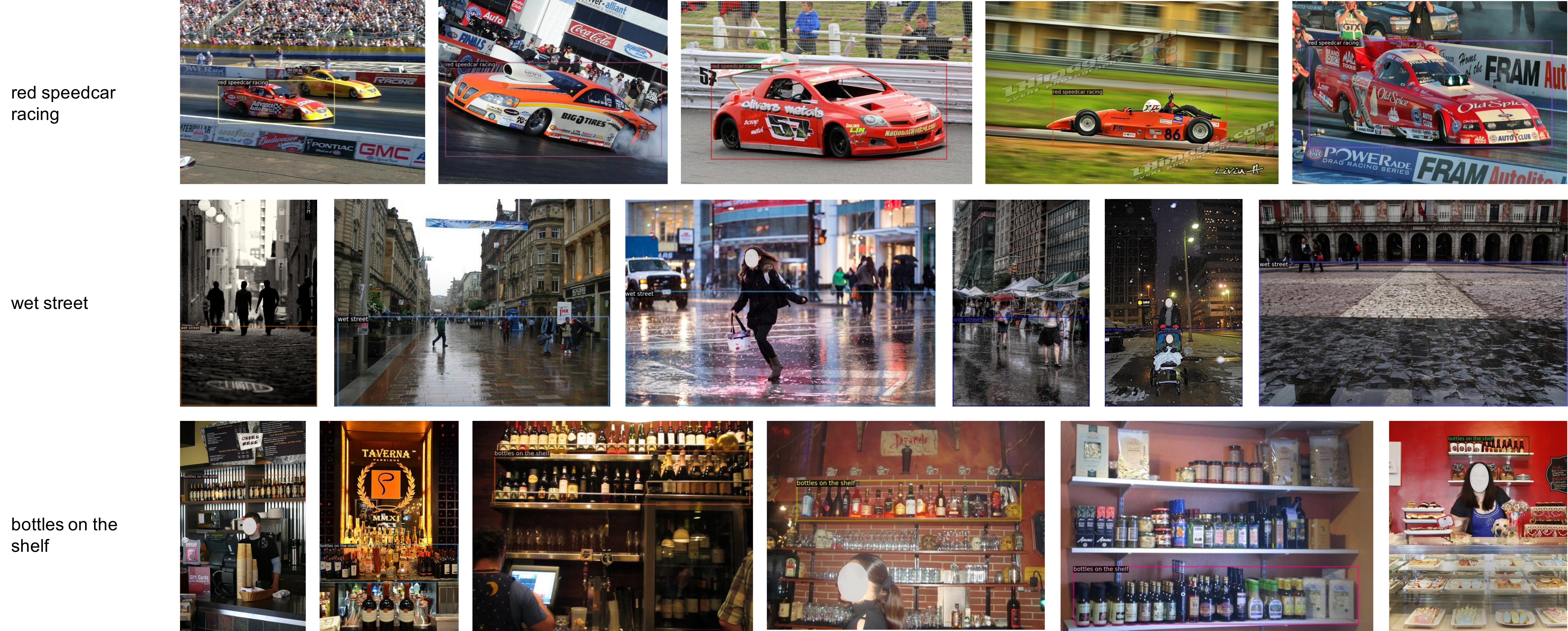}}\vspace{1mm}
\end{minipage}\\
\begin{minipage}[b]{1.\linewidth}
\centering
\centerline{\includegraphics[width=1.0\linewidth]{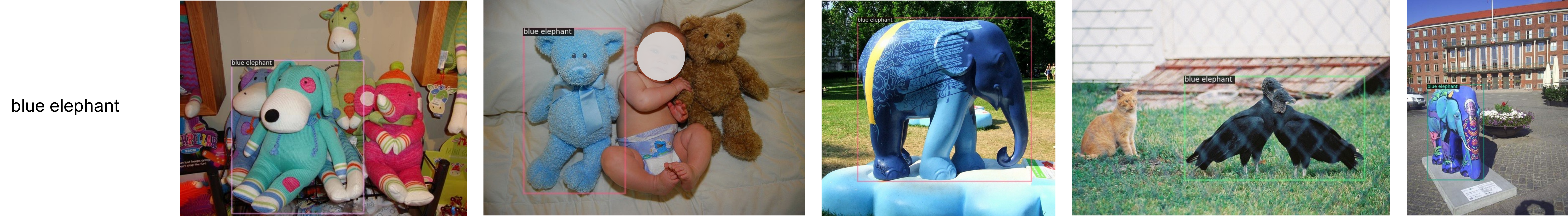}}
\end{minipage}
\caption{The top MMIS retrieval results for free-form language queries on the database of Objects365.}
\label{fig:mmis}\vspace{-3mm}
\end{figure*}

%=======================================================================
\section{Dataset Details}

We used the \emph{mixed} dataset of MDETR, which is a combination of Flicker30k entities \cite{DBLP:journals/ijcv/PlummerWCCHL17}, RefCOCO/RefCOCO+/RefCOCOg \cite{DBLP:conf/eccv/YuPYBB16,DBLP:conf/cvpr/MaoHTCY016}, Visual Genome (VG) \cite{DBLP:journals/ijcv/KrishnaZGJHKCKL17}, and GQA \cite{hudson2019gqa}. A typical example is shown in Figure \ref{fig:examples} (a). It can be found the query is a paragraph of queries and some objects are not annotated. As mentioned in the paper, we at first split the text query to a list of independent queries, as shown in Figure \ref{fig:examples} (b). Next, we append the COCO \cite{DBLP:conf/eccv/LinMBHPRDZ14} bounding box annotation (without category information) to the image, as shown in Figure \ref{fig:examples} (c).

To obtain the pseudo labeled data on LocNar \cite{DBLP:conf/eccv/Pont-TusetUCSF20}, given an image and its corresponding query, at first we use Spacy\footnote{https://spacy.io/} to extract the noun phrases which are possible objects in the text query.
Then we treat the pseudo-labeling as a phrase grounding task, retrieving the bounding box that is most aligned with the noun phrase. The model used for pseudo-labeling is trained on ``+CC'' of Table \ref{tab:ablation} in the paper. Some pseudo labeled examples are shown in Figure \ref{fig:locnar} (top row). It can be found that the pseudo labels are reasonably good. But they could be not accurate, especially when there are multiple objects present in the image for a single noun phrase. For example, in Figure \ref{fig:locnar} (a) (top row), we can only localize a single food item but miss the others because we do not know how many food items in this image. In addition, the OpenImages \cite{krasin2017openimages} object annotations (without category information) were also added to LocNar similar to COCO, as shown in Figure \ref{fig:locnar} (bottom row).

%=======================================================================
\section{MMIS Visualization}

We have shown more examples of MMIS retrieval results in Figure \ref{fig:mmis}. The database we used here is the Objects365 dataset \cite{shao2019objects365}. It has shown that X-DETR can accurately retrieve the most relevant instance, with bounding box, to the query. The model can discriminate the differences between objects with different attributes. For example, the query of ``skier in the sky'' finds skiers jumping in the sky instead of standing on the snow. And ``flying seagull'' and ``standing seagull'' find seagull flying and standing, respectively. When given ``red speedcar'', the retrieved results are common red speedcars, e.g., in the parking lot or building. But when the attribute of ``racing'' is added, the most relevant results are speedcars in racing games. These have shown the power of X-DETR for any free-form language description. However, MMIS is still a very challenging task, and some of the top retrieved results may not be correct. For example, the third result for ``skier in the sky'' is mistaken due to the camera angle, and last result for ``wet street'' is wrong because of the building shadow. Also for ``blue elephant'', the top two results are wrong, probably because the model has never seen blue elephants during training, and the database may not have any true examples of ``blue elephant''. But interestingly, X-DETR does find two blue elephant statues, which could be the most relevant results in the database. MMIS is different from image-text retrieval, where the target of interest is at object-level instead of image-level. For example, the last image for ``bottles on the shelf'' is unlikely be retrieved by image-text retrieval, because those bottles only occupy a small portion of the whole image.

%%%%%%%%% REFERENCES
{\small
\bibliographystyle{ieee_fullname}
\bibliography{egbib}
}

\end{document}